 \documentclass[sigconf]{acmart}
\acmSubmissionID{261}
\usepackage{booktabs} 
\usepackage{multirow}
\usepackage{tabularx}
\usepackage{wrapfig}
\usepackage{mathtools}
\usepackage{lipsum}
\usepackage{footmisc}
\usepackage{bbm}
\usepackage[ruled]{algorithm2e} 

\newcommand{\framework}{PACO}
\SetAlFnt{\small}
\SetAlCapFnt{\small}
\SetAlCapNameFnt{\small}
\SetAlCapHSkip{0pt}

\copyrightyear{2024}
\acmYear{2024}
\setcopyright{acmlicensed}\acmConference[SIGGRAPH Conference Papers '24]{Special Interest Group on Computer Graphics and Interactive Techniques Conference Conference Papers '24}{July 27-August 1, 2024}{Denver, CO, USA}
\acmBooktitle{Special Interest Group on Computer Graphics and Interactive Techniques Conference Conference Papers '24 (SIGGRAPH Conference Papers '24), July 27-August 1, 2024, Denver, CO, USA}
\acmDOI{10.1145/3641519.3657409}
\acmISBN{979-8-4007-0525-0/24/07}

\begin{document}
\def\negativevspace{}	
\newcommand{\todo}[1]{{\color{red}{[TODO: #1]}}}
\newcommand{\TODO}[1]{{\color{red}{[TODO: #1]}}}
\newcommand{\rh}[1]{{\color{green}#1}}
\newcommand{\dc}[1]{{\color{red}#1}}
\newcommand{\lzz}[1]{{\color{blue}#1}}
\newcommand{\new}[1]{{#1}}
\newcommand{\phil}[1]{{\color[rgb]{0.2,0.6,0.2}#1}}
\newcommand{\jy}[1]{{\color[rgb]{0.7,0.2,0.0}#1}}
\newcommand{\edward}[1]{{\color[rgb]{0.7,0.2,0.7}#1}}
\newcommand{\ednote}[1]{{\color[rgb]{0.7,0.2,0.7}ED: #1}}
\newcommand{\para}[1]{\vspace{.05in}\noindent\textbf{#1}}
\newcommand{\xjqi}[1]{{\color{magenta}#1}}
\def\ie{\emph{i.e.}}
\def\eg{\emph{e.g.}}
\def\etal{{\em et al.}}
\def\etc{{\em etc.}}
\newcolumntype{C}[1]{>{\centering\arraybackslash}p{#1}}
	

\title{Object-level Scene Deocclusion }

\author{Zhengzhe Liu}
\email{zzliu@cse.cuhk.edu.hk}
\affiliation{%
	\institution{The Chinese University of Hong Kong}\country{Hong Kong}} 
 \author{Qing Liu}
 \email{qingl@adobe.com}
 \affiliation{%
	\institution{Adobe}\country{United States}}
\author{Chirui Chang}
\email{u3010225@connect.hku.hk}
\affiliation{%
	\institution{The University of Hong Kong}\country{Hong Kong}}
 \author{Jianming Zhang}
 \email{jianmzha@adobe.com}
  \affiliation{%
	\institution{Adobe}\country{United States}}
 
 \author{Daniil Pakhomov}
 \email{dpakhomov@adobe.com}
  \affiliation{%
	\institution{Adobe}\country{United States}}
 \author{Haitian Zheng}
 \email{hazheng@adobe.com}
  \affiliation{%
	\institution{Adobe}\country{United States}}
 
 \author{Zhe Lin}
 \email{zlin@adobe.com}
 \affiliation{%
	\institution{Adobe}\country{United States}}
 
 \author{Daniel Cohen-Or}
 \email{cohenor@gmail.com}
\affiliation{%
	\institution{Tel Aviv University}\country{Israel}}
\author{Chi-Wing Fu}
\email{cwfu@cse.cuhk.edu.hk}
\affiliation{%
	\institution{The Chinese University of Hong Kong}\country{Hong Kong}
 }

\renewcommand\shortauthors{Liu et al.}

\begin{teaserfigure}
\vspace*{-1mm}
  \centerline{\includegraphics[width=0.99\textwidth]{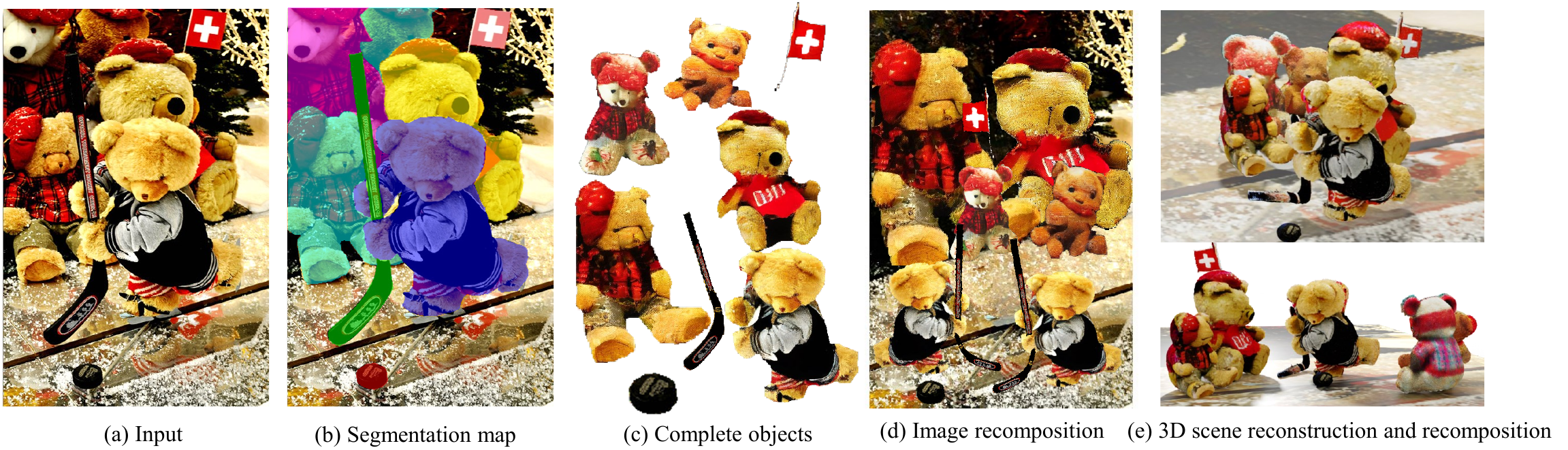}}
\vspace*{-3.5mm}
\caption{Given an image of a real-world scene (a), the image can be segmented into visible portions of the objects (b). Taking their category names in the dataset as text prompts,~\ie, ``teddy bear,'' ``flag,'' ``shaft,'' ``blade,'' and
``puck,'' the segments can be completed by our visible-to-complete latent generator that deoccludes the hidden portions (c).
This object-level deocclusion allows reconstruction and recomposition of the objects in the scene (d,e).
}
\label{fig:teaser}
\end{teaserfigure}

%
%


%
%



\begin{abstract}
%

Deoccluding the hidden portions of objects in a scene is a formidable task, particularly when addressing real-world scenes.
%
In this paper, we present a new self-supervised PArallel visible-to-COmplete diffusion framework, named \framework{}, a foundation model for object-level scene deocclusion. 
Leveraging the rich prior of pre-trained models, 
%
we first design the {\em parallel variational autoencoder\/}, which produces a full-view feature map that simultaneously encodes multiple complete objects, and the {\em visible-to-complete latent generator\/}, which learns to implicitly predict the full-view feature map from partial-view feature map and text prompts extracted from the incomplete objects in the input image. 
To train \framework{}, we create a large-scale dataset with 500k samples to enable self-supervised learning, avoiding tedious annotations of the amodal masks and occluded regions.
At inference, we devise a layer-wise deocclusion strategy to improve efficiency while maintaining the deocclusion quality. 
%
%
Extensive experiments on COCOA and various real-world scenes demonstrate the superior capability of \framework{} for scene deocclusion, surpassing the state of the arts
by a large margin. Our
method can also be extended to cross-domain scenes and novel categories that are not covered by the training set. 
%
Further, we demonstrate the deocclusion applicability of \framework{} in single-view 3D scene reconstruction and object recomposition. Project page: \url{https://liuzhengzhe.github.io/Deocclude-Any-Object.github.io/}. 

\end{abstract}

\maketitle

\section{Introduction}

Photographs of real-world scenes often contain numerous objects, in which the occlusion among objects typically makes the photo editing very challenging.
Many techniques have been developed to reveal and restore the obscured portions of objects. \textcolor{black}{Specifically, the deocclusion task aims to predict the occluded shape and appearance of objects in a real-world scene, given the visible mask of each object.}
Yet, achieving comprehensive scene completion at the object level is an important pursuit, albeit an exceptionally challenging one.
This difficulty arises because such a task inherently requires a comprehensive knowledge of the real world to address the intricacies of diverse object types and occlusion patterns, among other factors.


Owing to the aforementioned challenges, current attempts at scene deocclusion primarily focus on toy/synthetic datasets~\cite{burgess2019monet,greff2019multi,engelcke2019genesis,francesco2020object,monnier2021unsupervised} 
or typical object categories~\cite{yan2019visualizing,zhou2021human,papadopoulos2019make}; hence, they have limited capabilities to handle general real-world scenes. The state-of-the-art approach on real-world scene deocclusion is SSSD~\cite{zhan2020self}, which formulates 
the scene deocclusion as a regression task, solved using a discriminative approach. This approach, however, limits the model's ability to generate new contents, resulting in less 
satisfactory outcomes, as illustrated in Figure~\ref{fig:zhan}.
The recent introduction of foundation models with {exceptional generative capability and their inherent rich knowledge opens up 
a new opportunity to advance the frontier research of scene deocclusion.



In this work, we present PACO, a novel PArallel object-level COmpletion framework, leveraging the prior knowledge of pre-trained foundation model for supporting object-level scene deocclusion.
%
Existing foundation models cannot be directly employed for the scene deocclusion task.
First, existing models are designed for completing missing regions, such as 
inpainting~\cite{rombach2022high,lugmayr2022repaint}, requiring a given mask of the occluded region, which is not available in the scene deocclusion task.
Further, these models often cannot generate contents that preserve or complete the original object. For example, the model can be confused with which object the missing region belongs to (see the left example in Figure~\ref{fig:inpainting_bear}) and may create a new object instead of completing occluded part of an existing object (see the right example in Figure~\ref{fig:inpainting_bear}). More details are provided in Section~\ref{sec:inpainting}.
Second, the naive approach of de-occluding objects one-by-one using a diffusion model is far from efficient, due to the significant computational demand of the denoising process and the number of objects to handle in a scene.

%
Here, we leverage the progress of foundation models and design the PACO framework for object-level scene deocclusion.
The problem's context is visually depicted in Figure~\ref{fig:teaser}. Given a photograph (a) containing multiple partially-occluded objects with segments (b) and category names, our model can deocclude each of the objects, and complete them all (c). Once these objects are complete, various image editing and 3D applications are enabled (d \& e).

Specifically, we introduce a parallel variational autoencoder that can encode a stack of objects into a full-view latent feature map, and the decoder recovers the specific object given its partial query mask. Then, we train the visible-to-complete latent generator to generate a full-view feature map from a partial-view feature map extracted from segmented visible objects. Further, to train the model to learn to deocclude objects, we design an object ensemble dataset to allow self-supervised training and encourage the model to preserve the identity of the original object when completing it. Finally, in the test phase, to effectively handle challenging heavy occlusion patterns and reduce interference of nearby objects in real-world scenes, we leverage the depth information to determine the occlusion relation among objects, separate a scene into multiple depth layers when needed, and simultaneously de-occlude multiple objects at the same depth layer in one unified denoising pass.

%
%

Extensive experiments on the real-world COCOA dataset~\cite{zhu2017semantic} demonstrate that our model, which is trained on our object ensemble dataset, can efficiently deocclude objects in many real-world scenes, surpassing the state of the arts in terms of amodal mask accuracy and complete content fidelity. 
Additional experiments on out-of-distribution images~\cite{zhou2017scene} and real-world scenes captured by ourselves illustrate the generalization capability of our model. 
Further, our approach allows multiple downstream applications, including scene-level single-image 3D reconstruction and object rearrangement in images and 3D scenes.

\begin{figure}
\centering
\includegraphics[width=0.99\columnwidth]{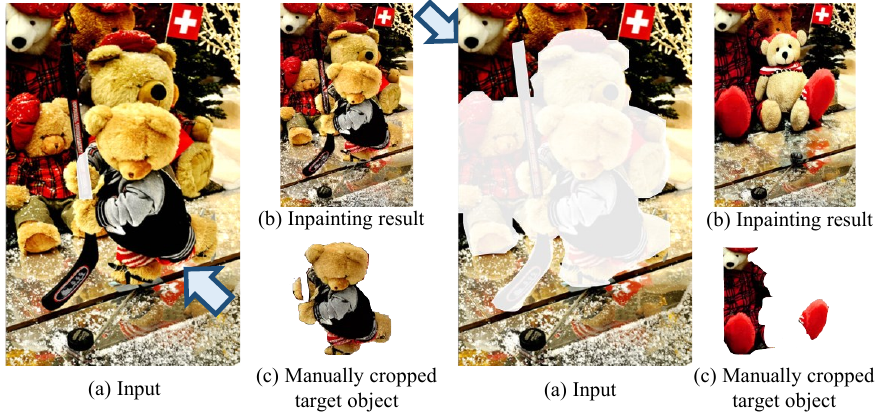}
\vspace*{-2.75mm}
\caption{Inpainting models, e.g., SD Inpainting~\cite{rombach2022high}, are not ready for the scene deocclusion task, even with the ground-truth amodal mask. The blue arrows mark the object to be deoccluded and the transparent white regions mark the missing areas to inpaint. Left: to inpaint the occluded region, the model can be confused with whether the missing region belongs to the occluder (hockey stick) or the occludee (front bear), failing to complete the target object (hand of front bear). Right: to inpaint regions behind all the occluders after removing them, the model may create unexpected new objects (a new bear) rather than deoccluding the target occludee behind.
}
\label{fig:inpainting_bear}
\vspace*{-2.75mm}
\end{figure}

\begin{figure*}
\centering
\includegraphics[width=0.99\textwidth]{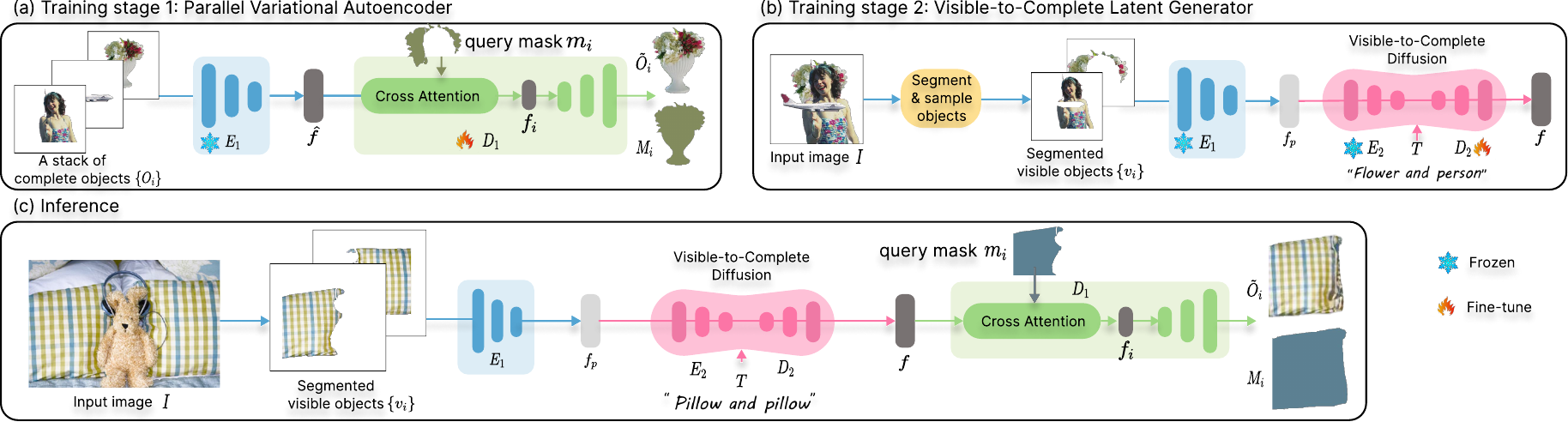}
\vspace*{-3.5mm}
\caption{Overview of our \framework{} framework.
(a) In the first training stage, we train the \textit{Parallel Variational Autoencoder} $\{E_1,D_1\}$ to learn to encode a stack of complete (full-view) objects $\{O_i\}$ into \textit{full-view} feature map $\hat{f}$ and the decoder $D_1$ to reconstruct the specific object $O_i$ for the \textit{partial query mask} $m_i$. 
(b) In the second training stage, we train the \textit{Visible-to-Complete Latent Generator} to generate full-view feature map $f$ conditioned on the partial-view features map $f_p$ from only segmented visible objects. 
(c) At inference, 
we employ the visible-to-complete latent generator to generate full-view feature map $f$ conditioned on partial-view feature map $f_p$ encoded from partial objects, then use $D_1$ to recover the amodal appearance $\tilde{O}_i$ with the partial mask $m_i$ as the query. 
}
\label{fig:overview}
\vspace*{-2.75mm}
\end{figure*}

\vspace*{-2.75mm}
\section{Related Work}
\label{sec:rw}


\paragraph{Image Inpainting}

A research area closely related to scene deocclusion is image inpainting. GANs~\cite{yu2019free,xie2019image} 
and diffusion models~\cite{rombach2022high,lugmayr2022repaint} to fill a missing image region, marked by a provided mask. 
Hence, it cannot be directly applied to scene deocclusion, as the inpainting models require a user-provided mask, which is not available in the deocclusion task. Also, even if the ground-truth mask of the occluded region is given, 
the completed contents may not be relevant to the occludee; illustrated in the left example of Figure~\ref{fig:inpainting_bear}. Further, 
it is not guaranteed that the generated contents preserve/complete the original contents, 
illustrated in the right example of Figure~\ref{fig:inpainting_bear}.

\vspace{-2mm}
\paragraph{Amodal Instance Segmentation}
Another research area related to scene deocclusion is amodal instance segmentation~\cite{zhu2017semantic,follmann2019learning,qi2019amodal,ke2021deep,purkait2019seeing,hu2019sail,xiao2021amodal,zhan2023amodal}, \textcolor{black}{which primarily targets the prediction of the occluded masks of the objects in a scene, without reconstructing their hidden appearance. }
Early works~\cite{kar2015amodal,li2016amodal} propose to predict the amodal bounding box and pixel-wise masks \textcolor{black}{that encompasses the entire extent of an object}, respectively. 
Other works~\cite{wang2020robust,yuan2021robust,sun2022amodal} integrate compositional models
for amodal segmentation.
Another line of works proposes 
semantic-aware distance maps~\cite{zhang2019learning}, amodal semantic segmentation maps~\cite{breitenstein2022amodal,mohan2022amodal,mohan2022perceiving}, and amodal scene layouts~\cite{mani2020monolayout,narasimhan2020seeing,liu2022weakly} for amodal prediction.

\vspace{-2mm}

\paragraph{Amodal Appearance Completion}

Scene deocclusion is also known as amodal appearance completion.
This task goes beyond amodal instance segmentation by not only predicting the occluded mask but also reconstructing the appearance of the occluded regions, making the task more complex and challenging. 
Early explorations~\cite{burgess2019monet,greff2019multi,engelcke2019genesis,francesco2020object,monnier2021unsupervised} typically work on toy 
datasets~\cite{greff2019multi,johnson2017clevr,Rishabh2019multi}. Due to the challenge of this task, some works attempt to complete the occluded appearance for 
specific
categories only,~\eg, 
vehicles~\cite{yan2019visualizing}, humans~\cite{zhou2021human}, and pizzas~\cite{papadopoulos2019make}. Due to the lack of training data with ground-truth occluded appearances, some works~\cite{ehsani2018segan,dhamo2019object,zheng2021visiting} try to predict the occluded appearance on synthetic datasets; yet, their performance on real-world scenes are 
not satisfactory, due to the synthetic-real domain gap. The state-of-the-art method is SSSD~\cite{zhan2020self}, which produces plausible results that clearly surpass the prior methods.
Yet, 
its generative fidelity is still far from satisfactory. 
\textcolor{black}{\cite{ozguroglu2024pix2gestalt}, a concurrent work with ours, can deocclude a single user-specified object but it lacks the efficiency to deocclude every object in the scene, unlike our approach.}

\vspace{-2mm}
\paragraph{Occlusion Order Prediction}
Some existing works~\cite{zhan2020self,nguyen2021weakly,ke2021deep,yuan2021robust} predict the occlusion order among objects in images. For example,~\cite{zhan2020self} pair-wisely predicts the occlusion order, yet requiring exhaustive multiple feed forwards.
\vspace{-2mm}
\paragraph{Diffusion Models}
Diffusion models~\cite{sohl2015deep} have shown impressive achievements on image generation~\cite{ho2020denoising,dhariwal2021diffusion}. 
Latent Diffusion Models ~\cite{rombach2022high} apply the diffusion process in the latent space instead of the pixel space to improve efficiency. In our work, we also adopt the latent diffusion approach, in which we encode a stack of complete objects into a full-view latent feature map and aim to generate it in a unified diffusion pass to improve efficiency. 



\section{Overview}

Given an image of a real-world scene, say $I$, we aim to deocclude the image,~\ie, to complete the occluded regions of the objects in the input image.
To do this, we design a novel self-supervised PArallel visible-to-COmplete diffusion framework named \framework{}.
%
%
Figure~\ref{fig:overview} gives an overview of the whole framework.


%
i)
In the first training stage, we create the {\em Parallel Variational Autoencoder\/} to (i) encode a stack of full-view (complete) objects $\{O_i\}$ into {\em full-view feature map\/} $\hat{f}$, which carries full-view information collectively of all the objects, and (ii) train decoder $D_1$ to {\em learn to recover specific object\/} for the given {\em partial query mask\/} $m_i$.
Full-view feature map $\hat{f}$ will be taken as the ground truth of feature map $f$ in the second training stage, whereas decoder $D_1$ will be employed to recover the full view of the specific object at inference.
%

ii)
In the second training stage, we train the {\em Visible-to-Complete Latent Generator\/} to learn to produce full-view feature map $f$ from partial-view feature map $f_p$.
As Figure~\ref{fig:overview} (b) shows, we first segment partially-visible objects $\{v_i\}$ from input image $I$ and encode them into the partial-view feature map $f_p$.
We can then train the visible-to-complete latent generator to learn to produce $f$ from $f_p$, meaning that we aim to perform the deocclusion implicitly in the latent space.
To support the training, we pre-encode the associated set of complete objects into full-view feature map $\hat{f}$ and take $\hat{f}$ as the target (ground truth) of feature map $f$.
Besides, we take each object's category name, provided by the dataset, to form a text prompt to condition the latent generator, such that we can reduce ambiguity.
%
%

iii)
With the trained visible-to-complete latent generator and decoder $D_1$, we are ready for inference.
Given an input image, we first segment it into partially-visible objects using SAM~\cite{kirillov2023segment}, then for each partial object, we create a partial mask 
and derive a text prompt,~\ie, the object's category name, using GPT-4V~\cite{openai2023gpt4v}.
After that, we can produce the partial-view feature map $f_p$ from the segmented partial objects and employ the visible-to-complete latent generator to produce the full-view feature map $f$ from $f_p$ together with the associated text prompt.
Last, we use decoder $D_1$ to recover specific full-view object $\tilde{O}_i$ from the generated feature map $f$ with the associated partial mask as a query.
%

In Section~\ref{sec:method}, we first present the two training stages,~\ie, the parallel variational autoencoder (Section~\ref{sec:ae}) and the visible-to-complete latent generator (Section~\ref{sec:v2a}).
After that, we present the layer-wise inference procedure (Section~\ref{sec:inference}) and describe how we prepare the dateset for training the framework (Section~\ref{sec:dp}).

\section{Methodology}
\label{sec:method}

\subsection{Parallel Variational Autoencoder}\label{sec:ae}
As illustrated in Figure~\ref{fig:overview} (a), from a stack of complete (full-view) object images $\{O_i\}$, we formulate our parallel variational autoencoder (VAE) to encode them into the full-view feature map $f\in\mathbb{R}^{H/r,W/r,c}$, such that $f$ carries full-view information of the objects, allowing recovery of any object in the stack.
Here, $H,W$ denote the resolution; $r$ is the downsampling rate; and $c$ is the channel number. The encoder $E_1$ extracts the feature of each input object image. These features are then aggregated by summing up the feature maps of all the object images, effectively avoiding potential bias towards any assumed order of the object images.
To optimize the memory usage, we employ an early-fusion technique, summing up the feature maps immediately after the initial convolution.
Afterward, the summed feature map is then mapped to a Gaussian distribution $\mathcal{N}(\mu,\sigma)$, where the full-view feature map $f$ can be sampled from.


%
Contrary to traditional VAEs, which focus typically on encoding a single image, our encoder $E_1$ is uniquely designed to encode multiple objects simultaneously. This parallel encoding capability enables the generation of multiple objects in a unified diffusion pass during the later stage of the visible-to-complete latent generator. Details of this process will be elaborated in Section~\ref{sec:v2a}.
%

Given full-view feature map $f$ and partial query mask $m_i$ associated with object $O_i$, decoder $D_1$ 
aims at reconstructing the complete full-view appearance of the object, $\tilde{O}_i$, with the associated amodal mask $M_i$.
Inside decoder $D_1$, we design a cross-attention~\cite{vaswani2017attention} mechanism to selectively extract the feature map $f_i$ associated with object $O_i$ from the full-view feature map $f$, utilizing the associated partial mask $m_i$ as the query: 
%
\begin{equation}
f_i=\text{softmax}(\frac{W_Q(m_i)W_K(f)^T}{\sqrt{c}}) W_V(f)
\end{equation}
where $W_Q$, $W_K$, and $W_V$ are convolutions to embed the inputs into query, key, and value, respectively. This process allows us to extract specific feature map $f_i$ from $f$, based on the given partial mask $m_i$, which then helps to reconstruct object $\tilde{O}_i$. This process is iteratively applied to each partial mask,~\ie, $m_1$, $m_2$, etc., to progressively deocclude all the objects in the input image stack. 

\paragraph{Discussion.}
The successful retrieval of individual feature map $f_i$ from the full-view feature map is primarily due to the cross-attention mechanism and the redundancy inherent in the feature representations. Specifically, selectively attending to relevant segments of the input feature map helps facilitate the target feature extraction in response to the query mask for $f$. 
Also, the network tends to learn redundant representations of the features. Even if some information is lost in the summation process, information about the object remains, so with the knowledge learned in the generative model, specific feature would eventually be reconstructed. Furthermore, to alleviate the representational load on $f$, we employ a layer-wise deocclusion strategy at the inference; see Section~\ref{sec:inference}.

On the other hand, it is noteworthy that we adopt the visible mask $m_i$ instead of the visible object $v_i$ as the query. The choice is based on our empirical findings that using $v_i$ as the query leads the decoder $D_1$ to complete object $o_i$ by overly relying on the appearance of $v_i$, rather than effectively leveraging the full-view feature map $f$. Such an approach tends to produce lower-quality reconstructions. Yet, using an appearance-absent query $m_i$, without the RGB information, encourages decoder $D_1$ to extensively exploit
$f$ in the reconstruction, yielding better image reconstruction quality.




\subsection{Visible-to-Complete Latent Generator}\label{sec:v2a}

\begin{figure}
\centering
\includegraphics[width=0.99\columnwidth]{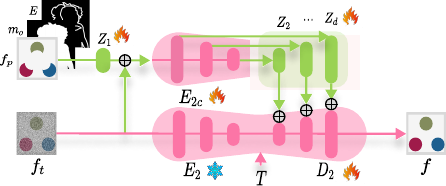}
\vspace*{-2.75mm}
\caption{Detailed architecture of our visible-to-complete latent generator.
}
\label{fig:v2c}
\vspace*{-2.75mm}
\end{figure}

The visible-to-complete latent generator is formulated as a U-Net architecture with a pair of encoder $E_2$ and decoder $D_2$, illustrated as a peanut shape in Figure~\ref{fig:overview} (b).
Given segments of partially-visible objects $\{v_i\}$ in image $I$ and texts of their category names $\{T_i\}$, our visible-to-complete latent generation aims to create the full-view feature map $f$ from just partial object images.
To do this, we first randomly sample a subset from $\{v_i\}$ as a data augmentation, encode the subset into a partial-view feature map $f_p$ using encoder $E_1$, then train a latent diffusion model to generate $f$ conditioned on $f_p$. 


The detailed architecture of our visible-to-complete latent generator is illustrated in Figure~\ref{fig:v2c}. 
We freeze the diffusion U-Net encoder $E_2$ 
to leverage the rich prior knowledge in the LDM foundation model~\cite{rombach2022high} and fine-tune $D_2$ to adapt the model to our deocclusion task. 
Inspired by ControlNet~\cite{zhang2023adding}, we clone $E_2$ to be a trainable copy $E_{2c}$, which takes $f_p$ as input. The cloning approach enables the network to incorporate $f_p$ as a conditional input without affecting the pre-trained encoder $E_2$. 
Following~\cite{zhang2023adding}, we insert zero convolutions $\mathcal{Z}_1,\cdots,\mathcal{Z}_d$ 
with weight and bias initialized as zeros, integrated at the entry point of $E_{2c}$ and throughout each layer of $D_2$. 
Consequently, the input of $E_{2c}$ is formulated as $f_t+\mathcal{Z}_1(f_p)$, so the initial layer input of $D_2$ can be expressed as $E_2(f_t)+\mathcal{Z}_2(E_{2c}(f_t+\mathcal{Z}_1(f_p)))$. Here, $f_t$ represents the noise-added variant of $f$ at diffusion timestamp $t$.
As our training commences, the initial inputs for both $E_{2c}$ and $D_2$ mirror those used during the LDM's pre-training phase, primarily because the zero convolutions ${\mathcal{Z}_1},\cdots,{\mathcal{Z}_d}$ are set to zeros initially. This setup not only aligns the starting conditions but also assists in minimizing the initial impact of random noise on the gradients. As a result, this approach facilitates a more effective transfer of the generative capabilities from the pre-trained model to $E_{2c}$ and $D_2$.

Besides $v_p$, we incorporate the mask $m_o$ of all sampled objects and their occluders, along with their edge map $E$ as conditions. These elements are resized to the same size as $v_p$ and then concatenated with $v_p$ to enhance the model's input conditions.
%
For text-based conditioning, we aggregate the category names of all sampled objects into a single text prompt $T$. This text prompt is then fed into the U-Net with cross-attention layers~\cite{rombach2022high}. The integration of textual information allows for a more nuanced and context-aware deocclusion process.


\subsection{Inference}\label{sec:inference}

\begin{figure}
\centering
\includegraphics[width=0.99\columnwidth]{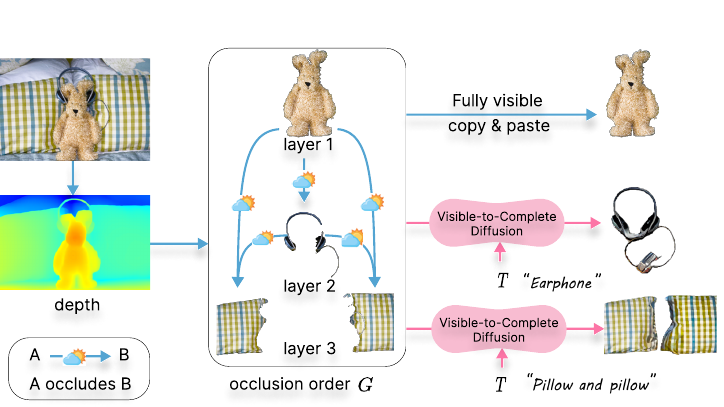}
\vspace*{-2.75mm}
\caption{Illustration of the layer-wise deocclusion strategy. Given an image, we first determine the occlusion relation among the objects using a depth estimation technique. Then, for each depth layer, we deocclude all objects in the same depth layer simultaneously in a unified diffusion pass.
}
\label{fig:occlusion}
\vspace*{-2.75mm}
\end{figure}

At inference, we utilize our model 
trained on our synthetic dataset to address the deocclusion of real-world scenes, as demonstrated in Figure~\ref{fig:overview} (c).
Initially, for a given real-world scene along with its segmentation map, as depicted in Figure~\ref{fig:occlusion}, we use a depth estimation model~\cite{ranftl2020towards} to assess the scene's depth. This allows us to determine the occlusion relation
for each pair of adjacent objects based on their relative depth across their shared boundary. Through this process, we can segment the scene into layers of object images.
All objects in the same depth layer can then be processed together through our parallel variational autoencoder, enabling deocclusion in a unified diffusion step. This layer-wise diffusion approach enhances the efficiency, 
while preserving high fidelity, as it also helps to reduce the potential interference among the objects, relieving the representative load of $f$.



For partially-visible objects $\{v_i\}$ in the same depth layer, we first apply our parallel VAE to encode them and yield a partial-view feature map $f_p$; see Figure~\ref{fig:overview} (c). Given $f_p$ and text condition $T$, namely the category names of the objects, we introduce random noise $f_N$ and employ our trained visible-to-complete latent generator to 
progressively refine the input feature map and generate $f$.
Subsequently, the partially-visible mask $m_i$ for each object $O_i$ is fed into $D_1$ to extract the associated object feature $f_i$ from $f$, such that we can then sequentially (layer by layer) generate the predicted full-view appearance $\tilde{O}_i$ and amodal mask $M_i$ of the object. 
Figure~\ref{fig:occlusion} demonstrates the layer-wise deocclusion strategy. Initially, we replicate the completely visible stuffed bunny as the uppermost depth layer. Subsequently, we employ our visible-to-complete latent generator to process the earphone in the second layer, and further process the two pillows in the third depth layer simultaneously.


Our parallel VAE architecture may support a ``Once-for-All'' strategy, where we encode all objects in the input image into $f$ and then deocclude all of them in a single diffusion denoising. While this approach significantly enhances efficiency, it has a high tendency of sacrificing the 
output image quality. 
The trade-off and effectiveness of this strategy will be further studied in Section~\ref{sec:ablation2}.
%
Also, to automate the preprocessing of our approach, we utilize SAM~\cite{kirillov2023segment} to generate the segment map and GPT-4V~\cite{openai2023gpt4v} to predict the object category names.
Further, while our approach focuses primarily on deocclusion at the object level, we can seamlessly leverage recent image inpainting models such as LAMA~\cite{suvorov2022resolution} 
to help us to deal with occluded background regions.




\subsection{Data Preparation for Self-Supervised Training}\label{sec:dp}

\paragraph{Dataset Creation}

To address the absence of a suitable dataset for the deocclusion task, we developed a new, extensive simulated Object Ensemble dataset (OE dataset) for training our model through self-supervision. We begin by utilizing a depth-guided method~\cite{ranftl2020towards} to estimate the occlusion order, allowing us to select $85k$ unobstructed objects from the COCO~\cite{lin2014microsoft} training dataset. These objects are then randomly arranged, one after another, on a blank canvas to create composite images, see,~\eg, the input image $I$ shown in Figure~\ref{fig:overview} (b). This simple yet effective approach helps to create a comprehensive dataset of $500k$ images that feature both occluded and complete objects.

In contrast to typical real-world image datasets, our OE dataset is specifically designed to encourage the model to concentrate on the object deocclusion task. It avoids the tendency of generating novel objects based on contextual information, a common characteristic of existing inpainting models like~\cite{rombach2022high}. The distinct advantages and outcomes of using the OE dataset over traditional approaches are highlighted in Figures~\ref{fig:inpainting_bear},~\ref{fig:inpaint_option} in the main paper, and Figure 1 in the supplementary file.


\paragraph{Training Strategy (Stage 1)}
Drawing inspiration from Latent Diffusion~\cite{rombach2022high}, our training framework is structured as a two-stage pipeline. In the first stage, to benefit from the extensive pre-existing knowledge in the pretrained LDM model~\cite{rombach2022high}, as Figure~\ref{fig:overview} illustrates, we keep encoder $E_1$ fixed and focus on optimizing decoder $D_1$ using the following loss functions.
%

(i) We use a regression loss to optimize the appearance prediction:
\begin{equation}
\label{eq:segment}
    \mathcal{L_{\text{r}}} = \sum_{i}\sum_{j}|| R_{i,j}-I_{i,j} ||_2^2,
\end{equation}
where $R_{i,j}$ and $I_{i,j}$ denote the reconstructed and ground-truth color of the $j$-th pixel in the $i$-th object image, respectively.

(ii) We adopt perceptual losses with the LPIPS metric~\cite{zhang2018unreasonable}, aiming to improve the visual fidelity of our reconstructions:
\begin{equation}
\label{eq:segment}
    \mathcal{L_{\text{p}}} = \text{LPIPS}(R,I).
\end{equation}

(iii) We incorporate an adversarial loss $L_{\text{adv}}$ in the first training stage, in which we optimize our decoder and a discriminator iteratively.
Following VAE~\cite{kingma2013auto,rezende2014stochastic}, we employ the Kullback-Leibler (KL) loss $L_{\text{kl}}$ to encourage the learned latent representation to follow a standard Gaussian distribution $\mathcal{N}(0,\mathbf{I})$ to facilitate efficient and effective learning.

(iv) For predicting the amodal mask $M_i$, we adopt a pixel-wise cross-entropy loss $\mathcal{L_{\text{m}}}$: 
\begin{equation}
\label{eq:segment}
    \mathcal{L_{\text{m}}} = -\sum_{i}\sum_{j}(\mathbbm{1}_{i,j}\log(p_{i,j}) + (1-\mathbbm{1}_{i,j})\log(1-p_{i,j})),
\end{equation}
where
$\mathbbm{1}_{i,j}$ is an indicator function (equals 1, if pixel $j$ is in object $i$) and $p_{i,j}$ is the predicted probability that pixel $j$ is in object $i$. 


The overall training loss for stage 1 is expressed as 
\begin{equation}
\label{eq:all}
    \mathcal{L} =  L_{\text{r}}+\lambda_1 L_{\text{p}}+\lambda_2 L_{\text{avd}}+\lambda_3 L_{\text{kl}}+\lambda_4 L_{\text{m}}.
\end{equation}
Here, the weighting factors $\lambda_1$ and $\lambda_3$ are determined in accordance to~\cite{rombach2022high} and $\lambda_4$ is set to be 0.3. 

\paragraph{Training Strategy (Stage 2)}
The visible-to-complete latent generator is designed to generate full-view feature map $f$ with the partial-view feature map $f_p$ as a condition.
%
In the forward diffusion process, noise is progressively added to feature map $f$ to transform it to a noisy sequence $\{f_t\}$, where $t$ is the time step.
Our model's objective is to reverse this process by learning to predict and remove the added noise $\epsilon$ at each time step.
Hence, an $L_2$ loss is employed:
%
%
\begin{equation}
\mathcal{L}_\text{v2c}=E_{t,C_0,\epsilon}[\left\| \epsilon-\epsilon_\theta(f_t,t,f_p,T,E,m_o \right\|^2], \epsilon\sim\mathcal{N}(0,\mathbf{I}),
\end{equation}
where 
$T$ denotes the text prompt condition, namely the category names of the encoded objects; and
$m_o$ and $E$ indicate the mask and edge map of the encoded objects and their occluders, as illustrated in Figure~\ref{fig:v2c}.
The function $\epsilon_\theta$ is the visible-to-complete U-Net for performing the denoising. Our model is trained to predict and reverse noise $\epsilon$ added at each time step, enabling an effective denoising and generation of the original feature map $f$.





\section{Experiments}

\subsection{Datasets, Implementation Details, and Metrics}

\framework{} is trained on a specially-created simulated object ensemble dataset, which comprises 500k images. Each image is assembled using two to eight object image samples from a pool of 85,000 objects taken from the COCO dataset~\cite{lin2014microsoft}. 

In the evaluation, we use the COCOA~\cite{zhu2017semantic} validation set (1,323 images) and test set (1,250 images). This work mainly focuses on object-level deocclusion, so we exclude those annotated as ``stuff'' in the dataset, such as the crowd, ice, etc. To demonstrate \framework{}'s ability to generalize across datasets, we test it also on ADE20k~\cite{zhou2017scene} and other novel scenes and categories. 

We build \framework{} using PyTorch. The model is initialized with weights from the Stable Diffusion model~\cite{rombach2022high}, then fine-tuned in the first training stage (parallel VAE) using the Adam optimizer with a learning rate of $4.5e^{-6}$. In the second training stage, the visible-to-complete latent generator is fine-tuned for 20k iterations with a learning rate of $1e^{-5}$ on 8 NVIDIA A100 GPUs. At inference, we use classifier-free guidance~\cite{ho2022classifier} with a scale factor of 9.



For quantitative evaluation on the quality of the amodal mask, we follow SSSD~\cite{zhan2020self}, to use the Intersection over Union (IoU) metric. This metric is calculated as $\text{IoU}=\frac{\sum_i{M_i\cap \hat{M_i}}}{\sum_i{M_i\cup \hat{M_i}}}$, where 
$i$ represents the object index, 
$M_i$ and $\hat{M}_i$ are the predicted and ground-truth amodal masks, respectively, and 
$\cup,\cap$ represent the union and intersection areas, respectively.
Also, we assess the image fidelity using the FID score~\cite{heusel2017gans}, comparing particularly the deoccluded objects with a ground-truth image set of unoccluded objects sourced from the COCOA training set. For a fair comparison, all object images are set against a white background.
Besides, we assess the accuracy of the occlusion order by evaluating pairwise accuracy on instances of the occluded object pairs, following SSSD.


\vspace{-2mm}
\subsection{Comparison with Existing Works}

\vspace{-1mm}
To validate the effectiveness of \framework{},  we conducted a comparative analysis with the state-of-the-art method,~\ie, SSSD~\cite{zhan2020self}. 
Table~\ref{tab:zhan} reports the quantitative comparison results, demonstrating the quality of \framework{} on both amodal mask accuracy (see ``IoU'' in Table~\ref{tab:zhan}) and fidelity of recovered appearance (see ``FID'' in Table~\ref{tab:zhan}). Additionally, the qualitative comparisons shown in Figure~\ref{fig:zhan} demonstrate that \framework{} can produce deocclusion results of higher fidelity. Notably, it is capable of reconstructing complete and plausible objects (like an elephant), generating high-quality textures consistent with visible regions (seen in the zebra and giraffes), maintaining true-to-life shapes (such as the rear wheel of the bus and the motor), and effectively deoccluding complex subjects like humans. 
These capabilities mark a significant advancement over the existing work.
%
Besides, our occlusion order prediction result also outperforms SSSD; see ``order accuracy'' in Table~\ref{tab:zhan}. \textcolor{black}{Note that both SSSD and our approach adopt a visible segmentation mask as input for deocclusion, which is a standard practice in this task. Besides, our model takes additional text descriptions, which are easy to acquire, either specified by users or using caption models, to better leverage the prior knowledge from the prior-trained foundation model.}



\begin{table}[t]
	\centering
		\caption{Quantitative comparison 
  on the COCOA validation set.
  }
        \vspace*{-2mm}
	\resizebox{0.9\linewidth}{!}{
		\begin{tabular}{C{2cm}|C{1.2cm}|C{1.2cm}|C{2.3cm}}
			\toprule[1pt]
                        Method &   IoU ($\uparrow$)    & FID  ($\downarrow$) & order accuracy ($\uparrow$) \\ \hline
                         SSSD  & 87.59 & 15.05 & 89.4 \\
                        \framework{} (ours) & \textbf{89.52}  & \textbf{13.93}  & \textbf{90.0} \\
			\bottomrule[1pt]
	\end{tabular}}
    \vspace*{-1mm}
\label{tab:zhan}
\end{table}
\begin{figure*}
\centering
\includegraphics[width=0.99\textwidth]{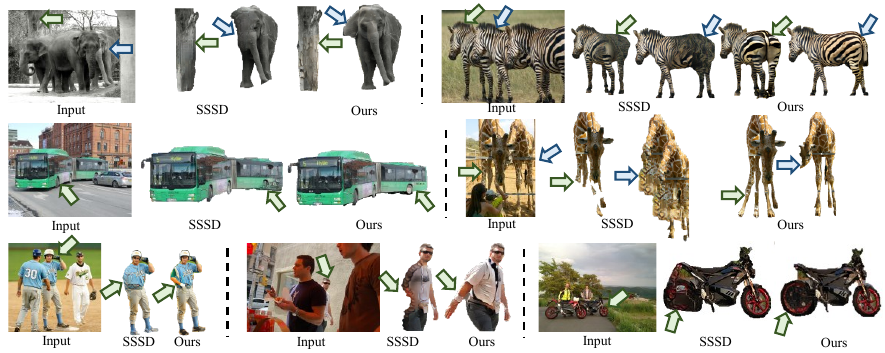}
\vspace*{-2.75mm}
\caption{Qualitative comparison with SSSD~\cite{zhan2020self}. The arrows indicate the target object to be deoccluded and the completed object parts. 
}
\label{fig:zhan}
\vspace*{-2.75mm}
\end{figure*}

Then we performed a qualitative comparison using the examples provided in their publication.
As Figure~\ref{fig:zheng} shows, the regions of the cup and car that were reconstructed by VINV exhibit noticeable blurring. This issue primarily arises from a significant domain gap between their synthetic data and real-world images, as well as the lower quality of their pseudo ground truths. In contrast, our method demonstrates a notable improvement in deoccluding objects like the cup, bread, and car, achieving a much higher fidelity level in the deoccluded objects. Please refer to the supplementary file for the quantitative comparison and more deocclusion results of our \framework{}. 


\begin{figure*}
\centering
\includegraphics[width=0.99\textwidth]{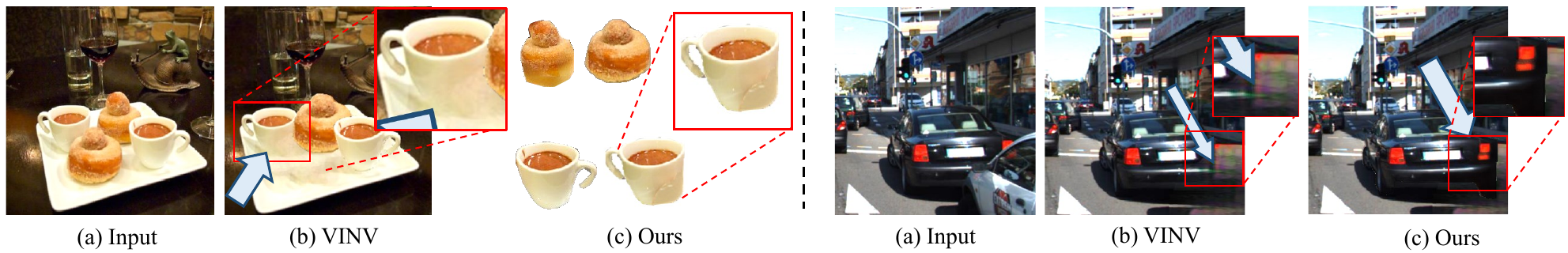}
\vspace*{-2.75mm}
\caption{Qualitative comparison with existing work VINV~\cite{zheng2021visiting}. Results of (b) are directly taken from their paper. The recovered regions from VINV, i.e., the bottom of the cup (left) and the tail light of the car (right), are blurry, while our approach gives higher-quality deocclusion results. 
}
\label{fig:zheng}
\vspace*{-2.75mm}
\end{figure*}

\begin{figure*}
\centering
\includegraphics[width=0.99\textwidth]{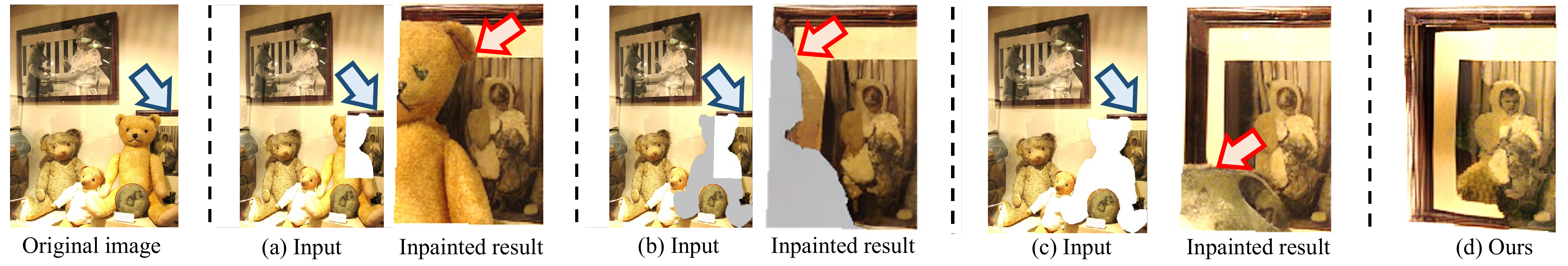}
\vspace*{-2.75mm}
\caption{Different inpainting strategies of Stable Diffusion inpainting~\cite{rombach2022high} along with GT amodal mask. (a) directly inpaints the ground-truth (GT) occluded region of the target object. (b) replaces the occluder (the teddy bear in this case) with a uniform color, specifically gray. (c) extends the inpainting mask to include the entire region of the occluder (the teddy bear). The white areas indicate the missing areas to inpaint. The blue arrows indicate the target object to be deoccluded. The red arrows means the unexpected objects generated by the inpainting model.   }
\label{fig:inpaint_option}
\vspace*{-1.75mm}
\end{figure*}

\vspace{-1mm}
\subsection{Comparison with Inpainting}~\label{sec:inpainting}

\vspace{-2mm}
We expanded our exploration to assess the feasibility of using existing inpainting models for scene deocclusion. Specifically, we experimented with three different strategies that apply the Stable-Diffusion inpainting.
Figure~\ref{fig:inpaint_option} shows the comparison results.
The first strategy (Figure~\ref{fig:inpaint_option} (a)) directly inpaints the ground-truth (GT) occluded region of the target object. This approach, however, leads to ambiguity, regarding which object the missing area belongs to. Hence, the inpainting model may mistakenly complete the occluding teddy bear instead of the intended album.
To address this issue, the second strategy (Figure~\ref{fig:inpaint_option} (b)) replaces the occluder (the teddy bear in this case) with a uniform color, specifically gray.  Despite this alteration, the inpainting process is partly influenced by the replacement color and the model still fails to accurately deocclude the album.
The third strategy (Figure~\ref{fig:inpaint_option} (c)) extends the inpainting mask to include the entire region of the occluder (the teddy bear).  However, this approach can lead the inpainting model to generate novel, unexpected objects in the occluded area.

In summary, all three strategies cannot successfully deocclude the album, even when using the ground-truth amodal mask, which is typically not accessible in real-world scenarios. In contrast, our approach, without an explicit amodal mask, can generate a complete and accurate representation of the album, as demonstrated in Figure~\ref{fig:inpaint_option} (d). This analysis highlights a crucial distinction between inpainting and deocclusion that they are fundamentally different tasks. Inpainting models, such as~\cite{rombach2022high}, are not inherently suited for deocclusion. \textcolor{black}{Note that our model requires only the ``visible masks,'' whereas the inpainting task requires the ``occluded invisible masks,'' which are much harder to derive.
}
%
Additional comparison results with GAN-based~\cite{suvorov2022resolution} and diffusion-based inpainting~\cite{rombach2022high} models along with GT amodal mask are included in the supplementary file.

\vspace{-1mm}

\subsection{Ablation Study}~\label{sec:ablation2}

\vspace{-2mm}

We study three alternative parallel deocclusion strategies.
(i) One-by-One serves as our baseline, as it deoccludes individual objects through successive diffusion processes. While it provides high-quality results, it is less efficient.
(ii) Layer-wise, as detailed in Section~\ref{sec:method}, encodes multiple objects in the same depth layer into a single feature map $f$.
(iii) Once-for-All, on the other hand, encodes all objects in a scene into the feature map $f$ and then takes a single diffusion process for scene deocclusion.

Quantitatively, as Table~\ref{tab:ablation2} reveals, the Layer-wise strategy attains a FID that is comparable to One-by-One, yet reducing the number of diffusion processes required per image by more than 65\% [(7.19-2.50)/7.19]. Furthermore, while Once-for-All boosts the efficiency significantly, it performs only one diffusion pass per scene, at the expense of the deocclusion fidelity.
Qualitatively, as Figure~\ref{fig:ablation2} depicts, Layer-wise and One-by-One are observed to deliver similar levels of deocclusion fidelity. While Once-for-All also yields credible results in simpler cases (see the left case in the figure), it exhibits a noticeable decline in performance in more challenging cases, such as with the zebras.
Indeed, the analysis of different deocclusion strategies illuminates the inherent trade-offs between efficiency and fidelity. 





\begin{table}[t]
	\centering
		\caption{Quantitative comparisons of different deocclusion strategies on the COCOA validation set. 
  }
        \vspace*{-2mm}
	\resizebox{0.8\linewidth}{!}{
		\begin{tabular}{C{1.9cm}|C{1cm}|C{4cm}}
			\toprule[1pt]
                        Strategy &   FID    &  Average number of diffusion processes per-image   \\ \hline
                         One-by-one  & \textbf{13.79} & 7.19 \\
                         Layer-wise & 13.93  & 2.50  \\
                         Once-for-all  & 14.56  & \textbf{1}  \\
			\bottomrule[1pt]
	\end{tabular}}
    \vspace*{-1mm}
\label{tab:ablation2}
\end{table}

\vspace{-1.5mm}

\subsection{Generalization Capability}

Leveraging the knowledge from pre-trained generative models, \framework{} can be extended to cross-domain scenes and novel categories not covered in our training dataset. This adaptability is evidenced by our results on the ADE20k dataset~\cite{zhou2017scene}, as showcased in Figure~\ref{fig:ade}. Here, our approach successfully deoccludes a wide variety of novel categories, including streams, plants, forklifts, and others.
Further, we evaluate our model on out-of-domain images, as illustrated in Figure~\ref{fig:out}. We introduce random occlusion regions in these images (Figure~\ref{fig:out} (b)) and apply our model to deocclude them. The outcomes (Figure~\ref{fig:out} (c)) demonstrate that our model adeptly handles the deocclusion of heavily obscured and unseen objects.

Also, \framework{}'s generalizability is further demonstrated through its performance on novel real-world scenes captured by our team. The examples shown in Figure~\ref{fig:our}, such as the fox and the toy horse, exhibit \framework{}'s robustness and versatility in diverse and unfamiliar scenarios, highlighting its strong generalization potential. 



\begin{figure*}
\centering
\includegraphics[width=0.99\textwidth]{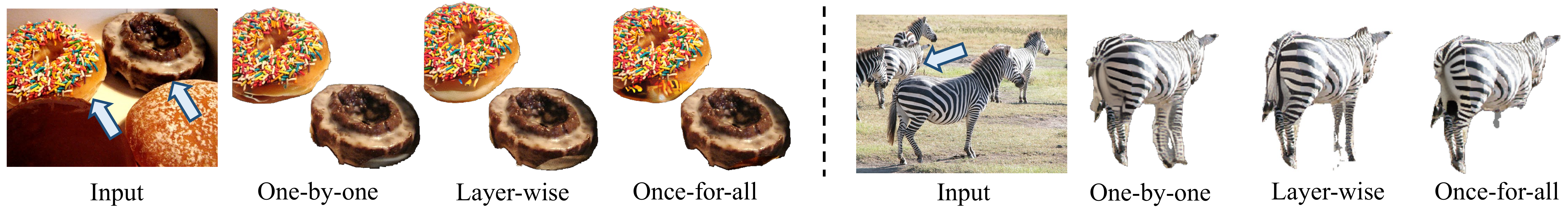}
\vspace*{-2.75mm}
\caption{Qualitative comparisons on deoccludion strategies. The blue arrows indicate the target object to be deoccluded. 
}
\label{fig:ablation2}
\vspace*{-1.75mm}
\end{figure*}

\begin{figure*}
\centering
\includegraphics[width=0.99\textwidth]{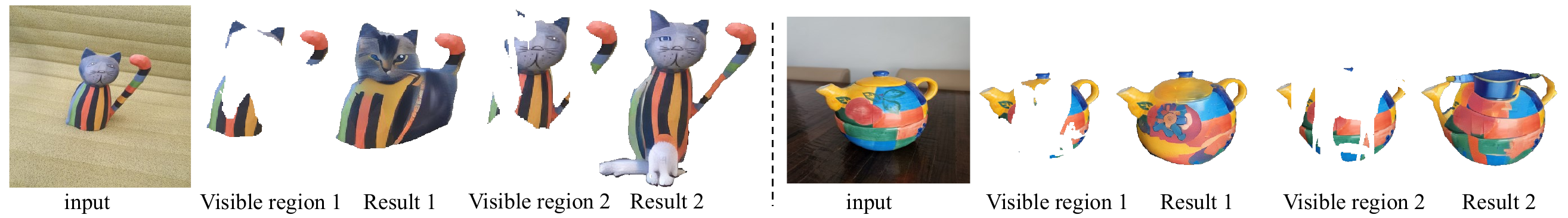}
\vspace*{-2.75mm}
\caption{Our deocclusion results on out-of-distribution images with random occlusion masks. The text prompts are ``a toy cat'' and ``a teapot''. 
}
\label{fig:out}
\vspace*{-2.75mm}
\end{figure*}

\begin{figure*}
\centering
\includegraphics[width=0.99\textwidth]{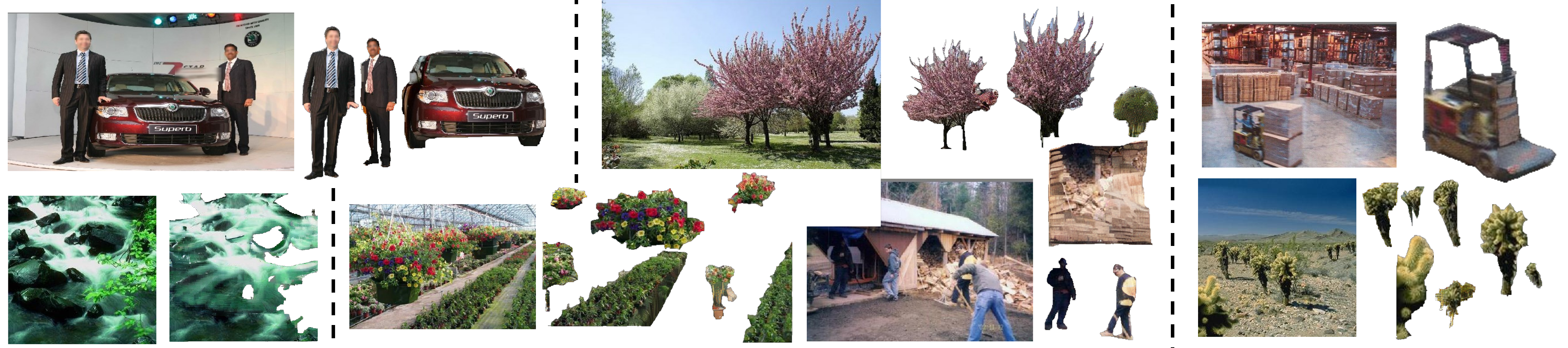}
\vspace*{-2.75mm}
\caption{Our deocclusion results on the ADE20k dataset~\cite{zhou2017scene}.
}
\label{fig:ade}
\vspace*{-2.75mm}
\end{figure*}

\begin{figure*}
\centering
\includegraphics[width=0.99\textwidth]{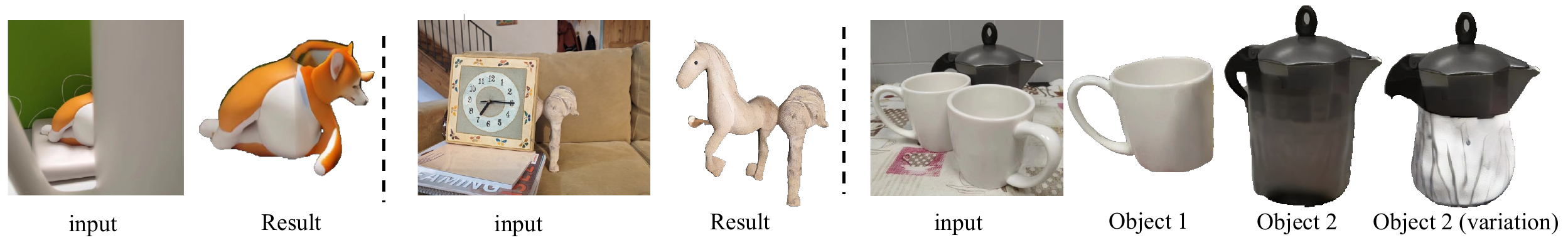}
\vspace*{-2.75mm}
\caption{Our results on real-world novel scenes captured by ourselves. The text prompts are ``a toy laying fox'', ``a toy horse'', ``a cup'', and ``a teapot''. 
}
\label{fig:our}
\vspace*{-2.75mm}
\end{figure*}

\vspace*{-2.75mm}

\begin{figure*}
\centering
\includegraphics[width=0.99\textwidth]{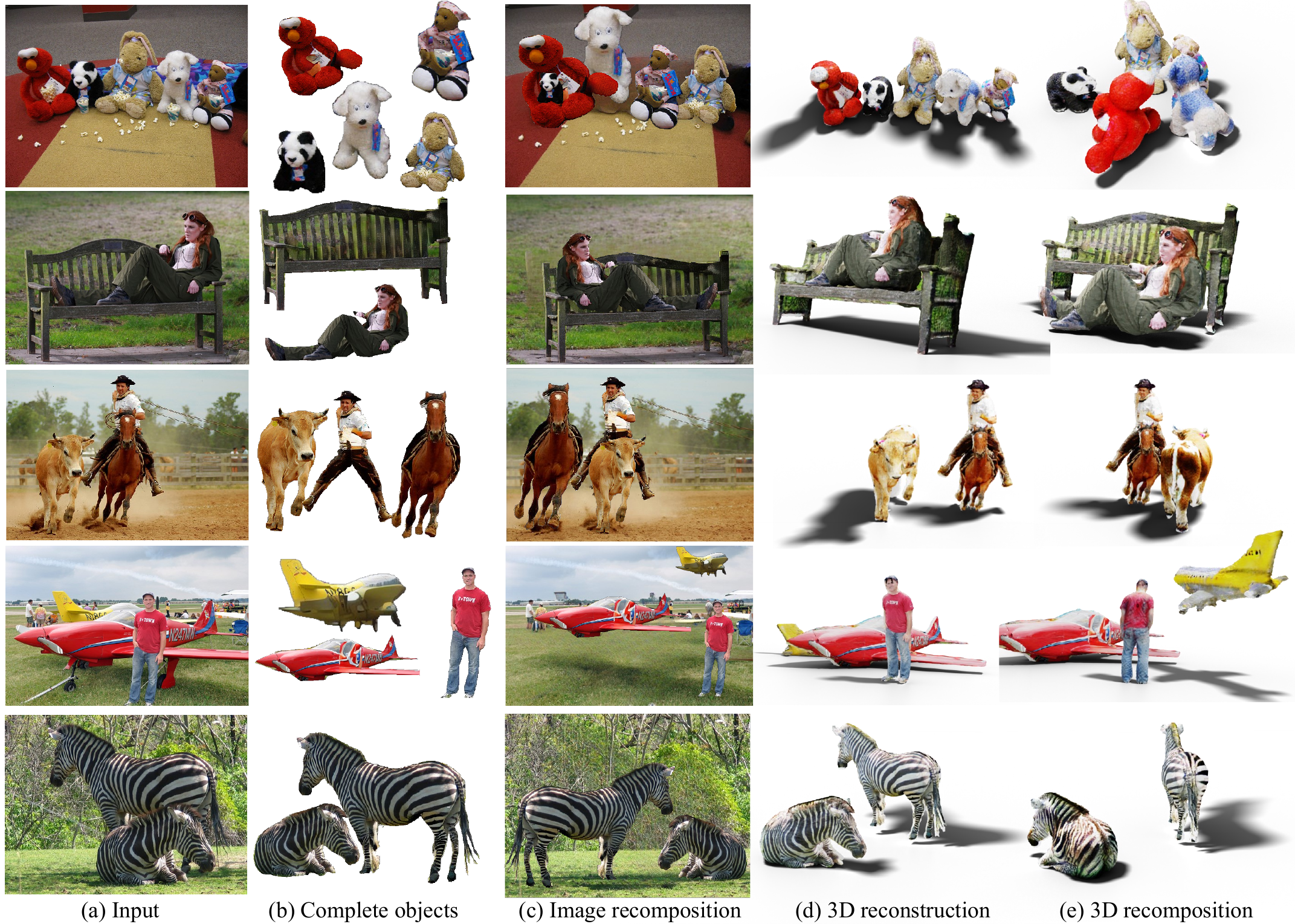}
\vspace*{-2.75mm}
\caption{Given an input image (a), our \framework{} deoccludes objects in it (b), enabling downstream applications including image recomposition (c), single-view 3D scene reconstruction (d), and 3D recomposition (e). 
}
\label{fig:3d}
\vspace*{-2.75mm}
\end{figure*}
\vspace{-3mm}
\subsection{Applications}

Figure~\ref{fig:3d} showcases the versatile applications made possible by our \framework{} framework. Our framework not only enables image recomposition but also extends to innovative domains such as single-view 3D scene reconstruction and 3D recomposition. Utilizing the deoccluded objects obtained through our framework and the backgrounds inpainted by models such as LAMA~\cite{suvorov2022resolution}, users gain the flexibility to interact with images, say to drag, resize, flip, and rotate objects, as well as alter their occlusion order, as demonstrated in Figure~\ref{fig:3d} (c). We also develop a GUI to allow users to easily conduct such operations; see the attached video. 

Moreover, our approach is compatible with recent advancements in object-level 3D reconstruction, such as DreamCraft3D~\cite{sun2023dreamcraft3d}, for novel applications, including the reconstruction of 3D scenes from a single viewpoint (Figure~\ref{fig:3d} (d)) and the recomposition of these 3D scenes (Figure~\ref{fig:3d} (e)). To do these, we first deocclude the objects and lift each object to 3D using~\cite{sun2023dreamcraft3d}. Then, we can assemble the 3D models based on their original locations in the scene or in other customized arrangements as desired. 
Such capabilities significantly expand the potential applications of our \framework{}, opening up new avenues for creative and practical usages in fields, ranging from digital content creation to mixed reality and beyond; see the illustrations in the attached video. 
Also, note that the existing 3D reconstruction approaches such as DreamCraft3D~\cite{sun2023dreamcraft3d} cannot be readily adopted for scene-level reconstruction; see the supplementary file for more details. 



\section{Conclusion and Limitations}

We introduced \framework{}, a method for image-space scene deocclusion. In presenting the deocclusion task, we distinguish it as a distinct challenge separate from the well-established inpainting task. We emphasize the significant differences between the two, highlighting that while both reveal unseen regions in an image, deocclusion is inherently object-aware, requiring object-level image understanding and completion of hidden parts associated with the occluded objects.
%

Our method leverages extensive knowledge of a pre-trained model, with fine-tuning specifically directed to the deocclusion task, preserving the model prior. However, several limitations are acknowledged. Challenges include addressing shadows and global lighting effectively and necessitating high-quality segment maps and text annotations per object, thus impeding automation. Also, the current implementation is not equipped to handle objects partially beyond the image boundary. Besides, due to the lack of full 3D awareness, it may occasionally yield unreasonable results. A more detailed exploration of these limitations is given in the supplemental material.

In our work, we facilitate self-supervised training by creating the synthetic object ensemble dataset, enhancing the preservation of the object identity. This self-supervision can scale, and as we look ahead, the training data can be further expanded, for enhanced performance, or possibly be reduced to be focus on a specific domain for efficiency.
Besides, we presented a number of applications based on deocclusion. We believe many more applications would require object-level deocclusion, and particularly, more research effort are needed to achieving fully automated deocclusion.



\begin{acks}
 This work is supported by Research Grants Council of
the Hong Kong Special Administrative Region, China (Project No.
CUHK 14201921).
\end{acks}

\vspace{+20mm}
\centerline{\Large{\textbf{Supplementary Material}}}

\section{Comparisons with Existing works and Inpainting + GT Amodal mask}

In this section, we conduct additional qualitative comparisons of our approach with: (i) SSSD~\cite{zhan2020self}, (ii) one of the most advanced GAN-based inpainting models LAMA~\cite{suvorov2022resolution} along with the GT amodal mask, and (iii) Stable Diffusion (SD) inpainting~\cite{rombach2022high} with GT amodal masks. These comparisons are illustrated in Figure\ref{fig:existing}. For these tests, we employed the third inpainting strategy indicated in Figure 8 (c) in the main paper, as our empirical findings indicated its superiority over the other two strategies in most scenarios.

As Figure~\ref{fig:existing} demonstrates, SSSD~\cite{zhan2020self} (Figure~\ref{fig:existing} b) and LAMA~\cite{suvorov2022resolution} with GT amodal mask (Figure~\ref{fig:existing} c) often results in blurry outcomes due to their limited generative capabilities. Stable Diffusion Inpainting~\cite{rombach2022high} with GT amodal mask (Figure~\ref{fig:existing} d) tends to utilize the context of the image to create new objects within the inpainting region, which diverges from the user's intention to deocclude the target object that is indicated by the blue arrow. In contrast, our method successfully generates results that are both high in fidelity and visually consistent with the original scene. Importantly, our approach maintains the identity of the original object without introducing any unintended new elements like Stable Diffusion inpainting; as indicated by the red arrows. In addition, it's crucial to note that inpainting models cannot be applied to deocclusion tasks without the availability of GT or user-defined amodal masks, 
highlighting the distinct nature and requirements of deocclusion compared to traditional inpainting tasks.

\begin{figure*}
\centering
\includegraphics[width=0.99\textwidth]{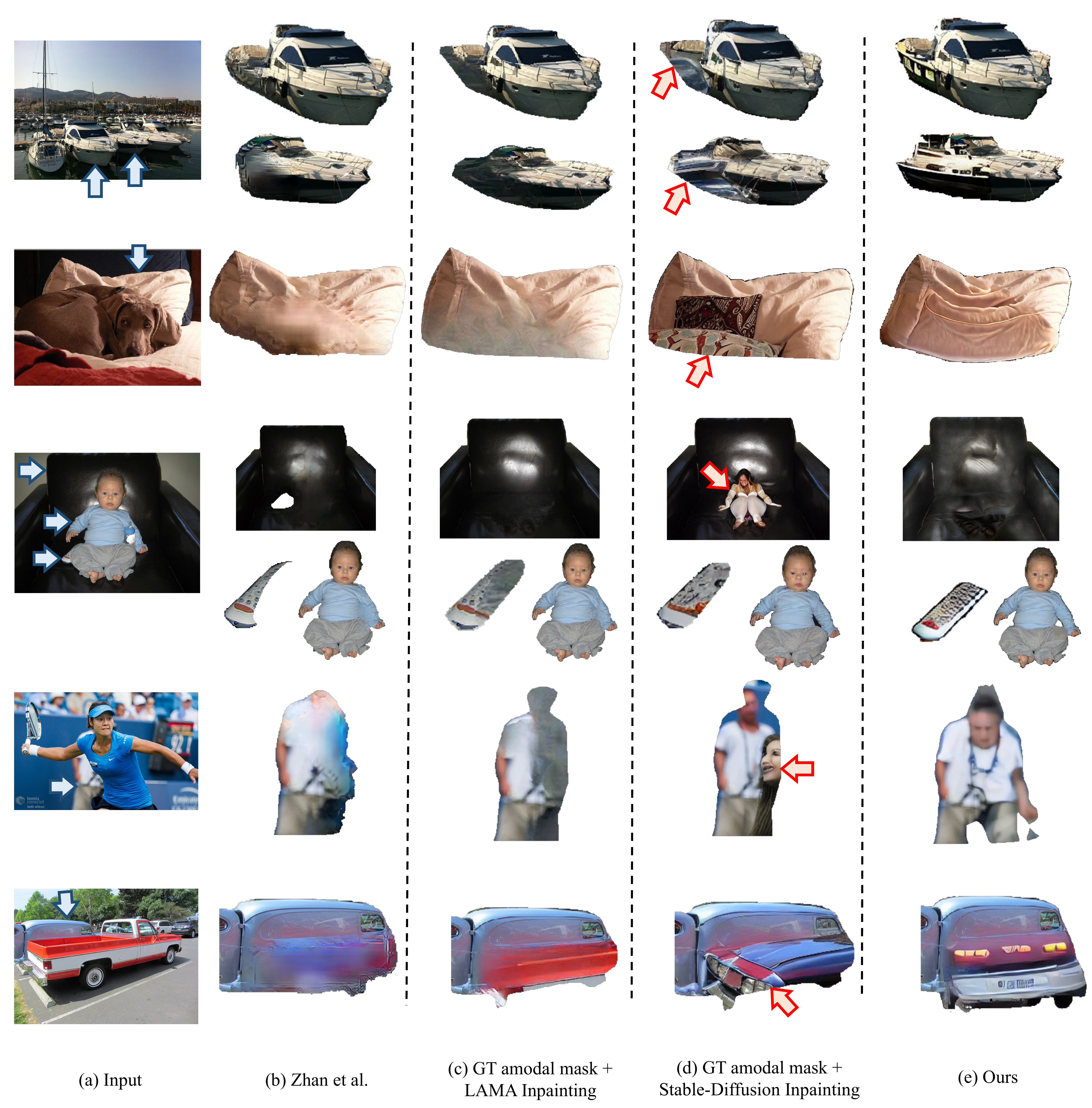}
\vspace*{-2.75mm}
\caption{Comparison with (b)~\cite{zhan2020self}, (c) LAMA~\cite{suvorov2022resolution} + GT amodal mask, and (d) SD inpainting~\cite{suvorov2022resolution} + GT amodal mask. The blue arrows indicate the target objects to be deoccluded. Note that the deoccluded objects in (d) are occluded by newly generated unexpected objects (indicated by the red arrows); thus, (d) cannot produce the desired deocclusion result even though the target object is manually cropped rather than using GT amodal mask.  
}
\label{fig:existing}
\vspace*{-2.75mm}
\end{figure*}

\section{Additional Results on Image Recomposition, 3D Reconstruction and Recomposition}

Complementing Figure 13 in the main paper, we showcase additional results of our downstream applications in Figure~\ref{fig:3d_supp}, including image recomposition, 3D reconstruction, and 3D recomposition. Moreover, Figure~\ref{fig:3d_supp} (f) displays outcomes obtained by directly applying the object-level 3D reconstruction method DreamCraft3D~\cite{sun2023dreamcraft3d} to multiple objects in a scene, without first deoccluding them using our method. However, this baseline approach leads to incomplete results, such as a partially reconstructed vase (top example in Figure~\ref{fig:3d_supp} (f)) and merged giraffes (bottom example in Figure~\ref{fig:3d_supp} (f)), illustrating the inadequacy of current object-level 3D reconstruction methods like DreamCraft3D for scene-level reconstruction.

In contrast, our PACO effectively decomposes a scene into complete, individual objects. This capability enables PACO to extend object-level 3D reconstruction approaches to innovative applications, such as scene-level 3D reconstruction and recomposition. 


\begin{figure*}
\centering
\includegraphics[width=0.99\textwidth]{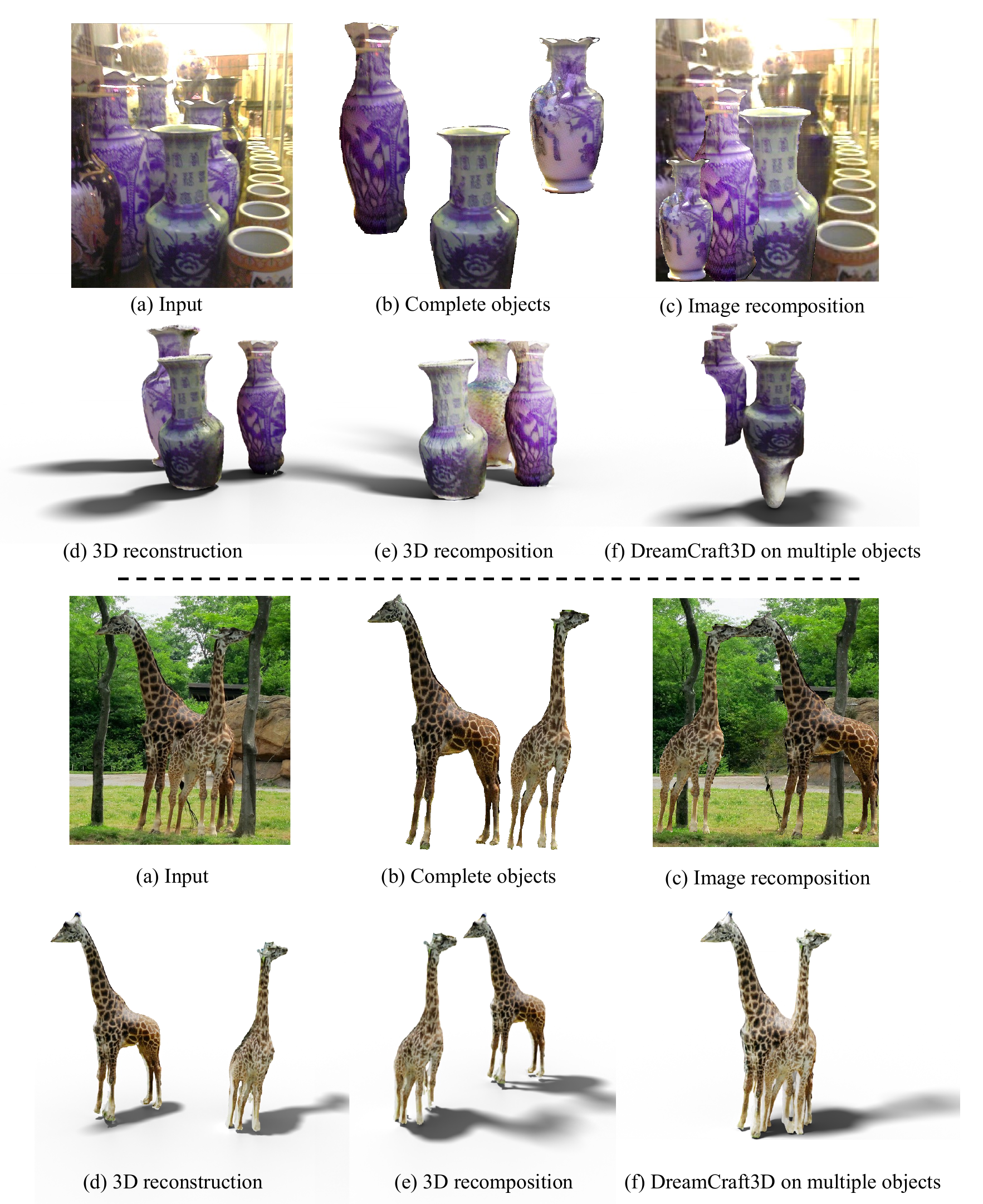}
\vspace*{-2.75mm}
\caption{Additional results on (b) object deocclusion, (c) image recomposition, (d) 3D reconstruction, and (e) 3D recomposition. (f) is derived by directly applying the object-level 3D reconstruction approach DreamCraft3D~\cite{sun2023dreamcraft3d} on multiple objects in the input image without using our deocclusion method. This approach resulted in incomplete reconstructions, such as partially formed objects (top example), and incorrectly merged objects, like the conjoined giraffes in the bottom example. }
\label{fig:3d_supp}
\vspace*{-2.75mm}
\end{figure*}

\section{Evaluation on our Object Ensemble Dataset}

To assess the reconstruction capabilities of our parallel VAE, we evaluate it using our object ensemble dataset. This dataset is particularly useful since real-world datasets lack ground truth of the occluded appearances. Additionally, we evaluate our visible-to-complete latent generator on this synthetic dataset, which eliminates the domain gap between the training and testing images. For this evaluation, we have set aside a validation set created through our data generation pipeline.

As depicted in Figure~\ref{fig:ablation1} (c), our parallel VAE demonstrates an impressive ability to accurately reconstruct the objects present in the input image (a). Furthermore, in (d), our visible-to-complete latent generator shows superior deocclusion capability on our object ensemble dataset, especially for the challenging cases where the objects are heavily occluded; see the two deoccluded persons in the top and bottom examples in Figure~\ref{fig:ablation1} (d). 

The quantitative results of these evaluations, presented in Table~\ref{tab:ablation1}, again underscore the exceptional reconstruction capabilities of our parallel VAE. Moreover, these results suggest that the generative quality of our visible-to-complete latent generator has the potential for further enhancement without the simulation-to-reality (sim-to-real) gap. 


\begin{table}[t]
	\centering
		\caption{Quantitative results from our stage-1 parallel VAE and stage-2 visible-to-complete latent generator on our synthetic object ensemble validation set. 
  }
        \vspace*{2mm}
	\resizebox{0.7\linewidth}{!}{
		\begin{tabular}{ccc}
			\toprule[1pt]
                        Method &   Stage-1 (Parallel VAE) &  Stage-2 (V2C Diffuison)     \\ \hline
                         FID  & 10.68  & 13.28   \\
			\bottomrule[1pt]
	\end{tabular}}
    \vspace*{-1mm}
\label{tab:ablation1}
\end{table}

\begin{figure*}
\centering
\includegraphics[width=0.99\textwidth]{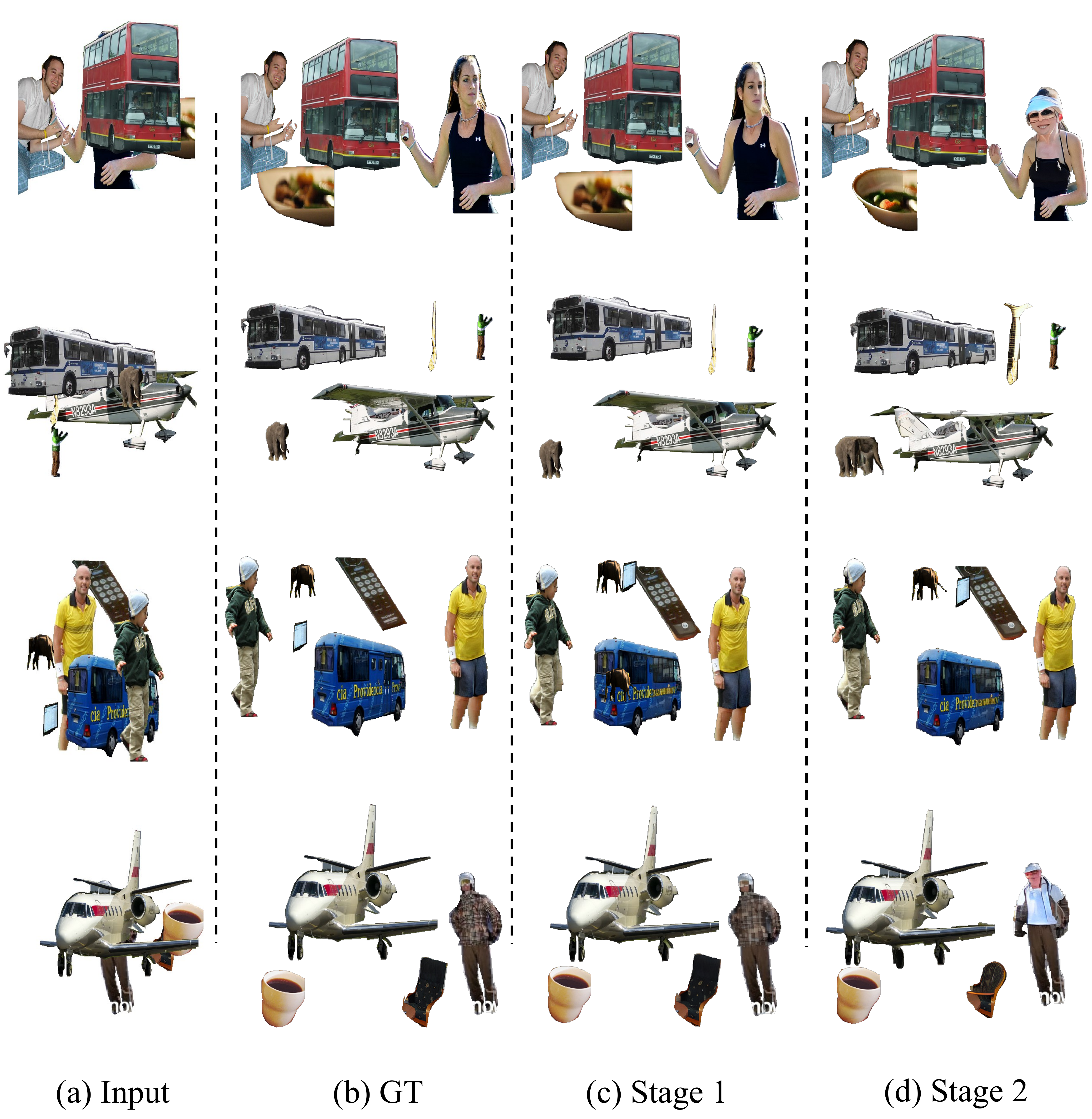}
\vspace*{-2.75mm}
\caption{Results of our stage-1 parallel VAE (c) and stage-2 visible-to-complete latent generator (d) on our synthetic object ensemble dataset. }
\label{fig:ablation1}
\vspace*{-2.75mm}
\end{figure*}

\section{Additional Deocclusion Results}

We show more deocclusion results from texts in Figures~\ref{fig:1},~\ref{fig:2},~\ref{fig:3},~\ref{fig:4},~\ref{fig:5},~\ref{fig:6},~\ref{fig:7}, and~\ref{fig:8}. These results further exemplify the effectiveness of our PACO in handling a broad spectrum of object categories across various real-world scenes.

\begin{figure*}
\centering
\includegraphics[width=0.99\textwidth]{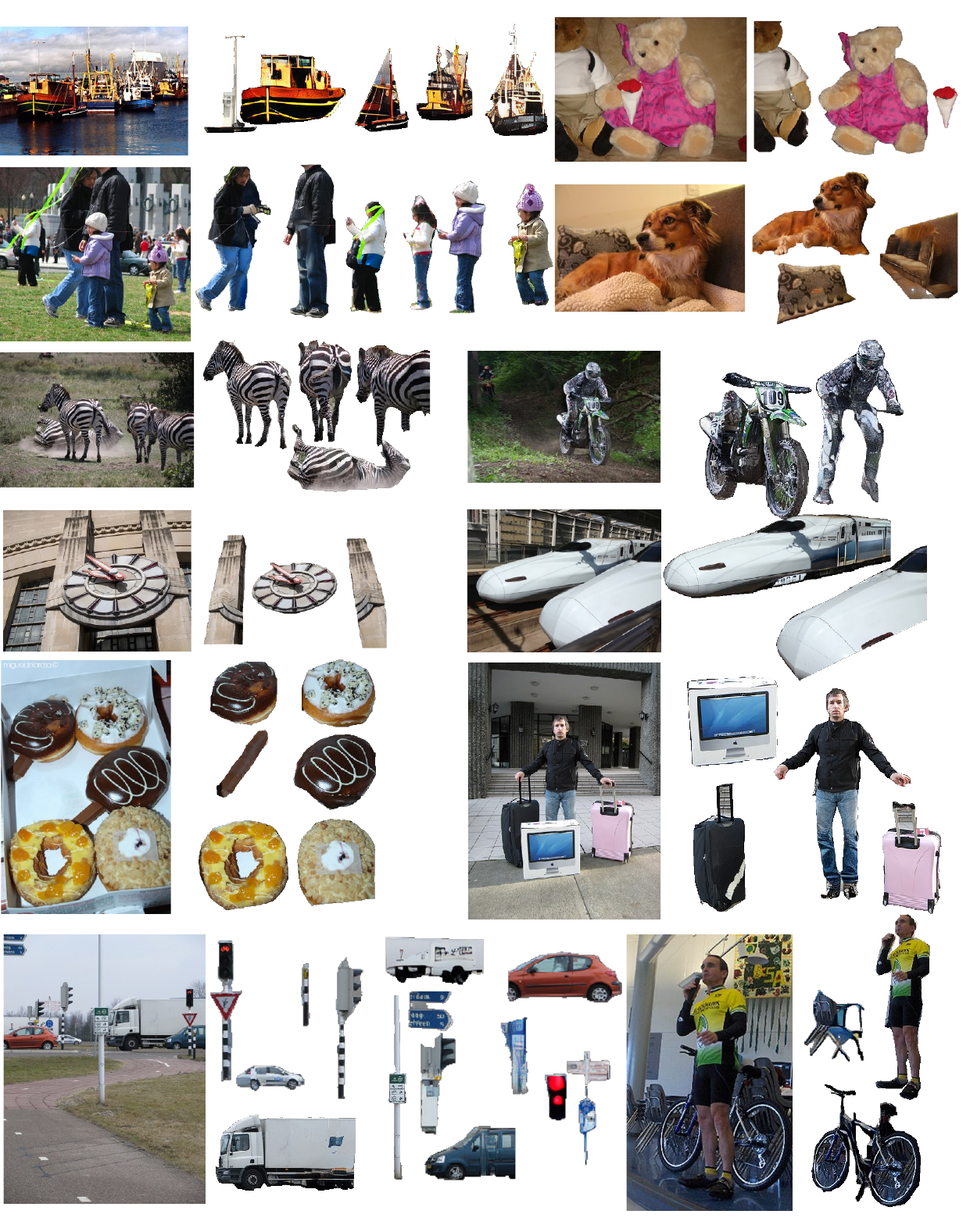}
\caption{Additional deocclusion results of PACO in real-world scenes. 
}
\label{fig:1}
\vspace*{-2.75mm}
\end{figure*}

\begin{figure*}
\centering
\includegraphics[width=0.99\textwidth]{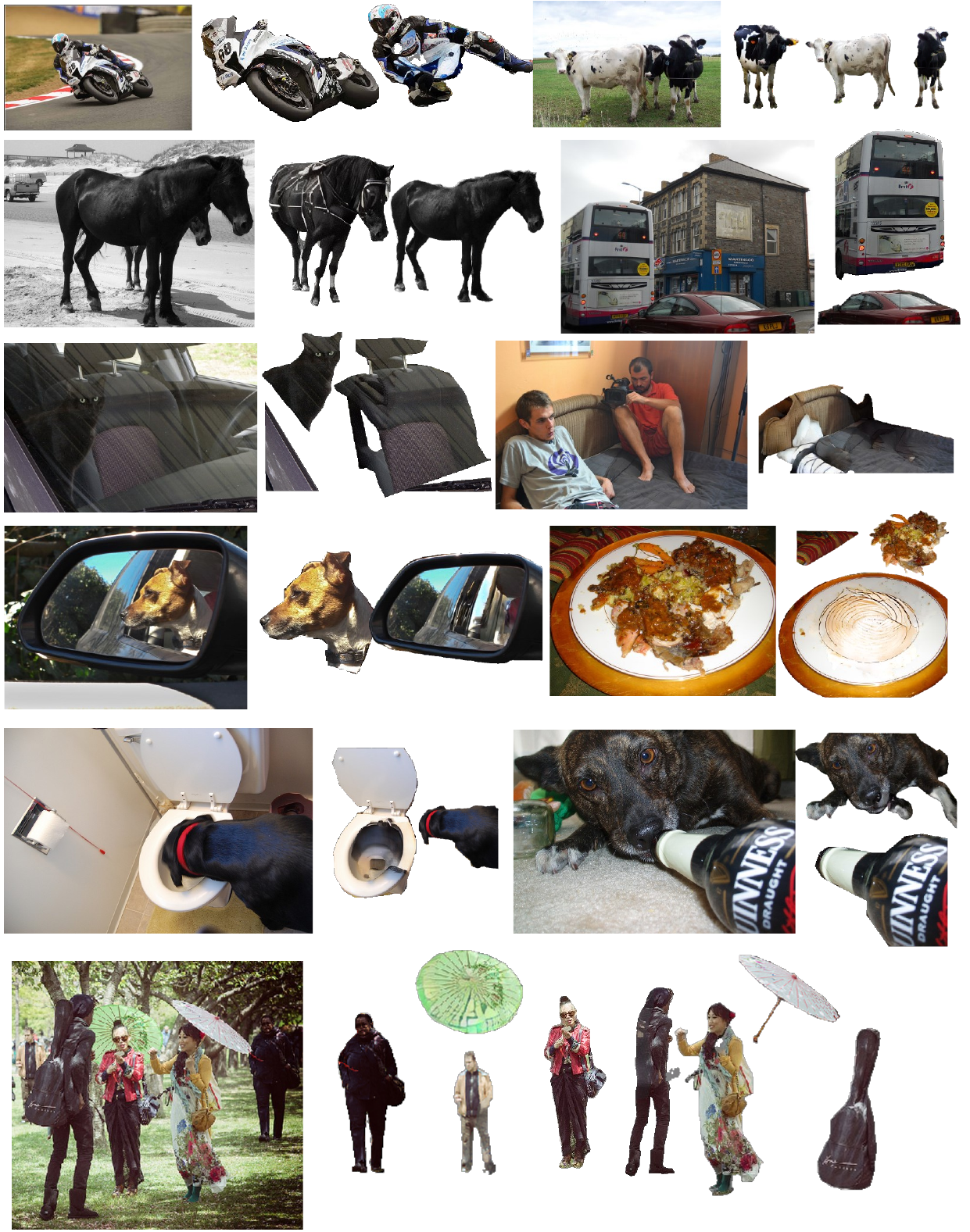}
\caption{Additional deocclusion results of PACO in real-world scenes. 
}
\label{fig:2}
\vspace*{-2.75mm}
\end{figure*}

\begin{figure*}
\centering
\includegraphics[width=0.99\textwidth]{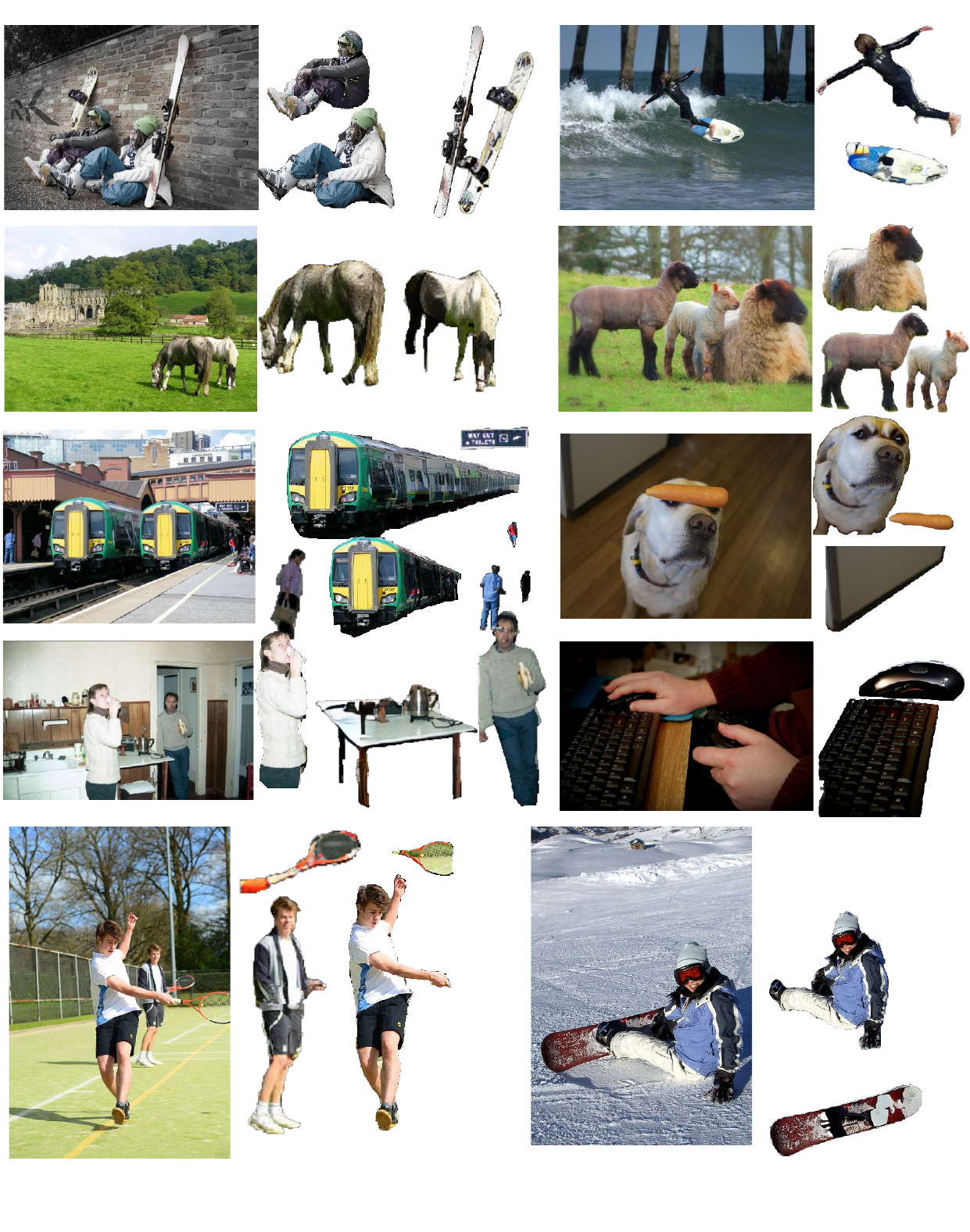}
\caption{Additional deocclusion results of PACO in real-world scenes.
}
\label{fig:3}
\vspace*{-2.75mm}
\end{figure*}

\begin{figure*}
\centering
\includegraphics[width=0.99\textwidth]{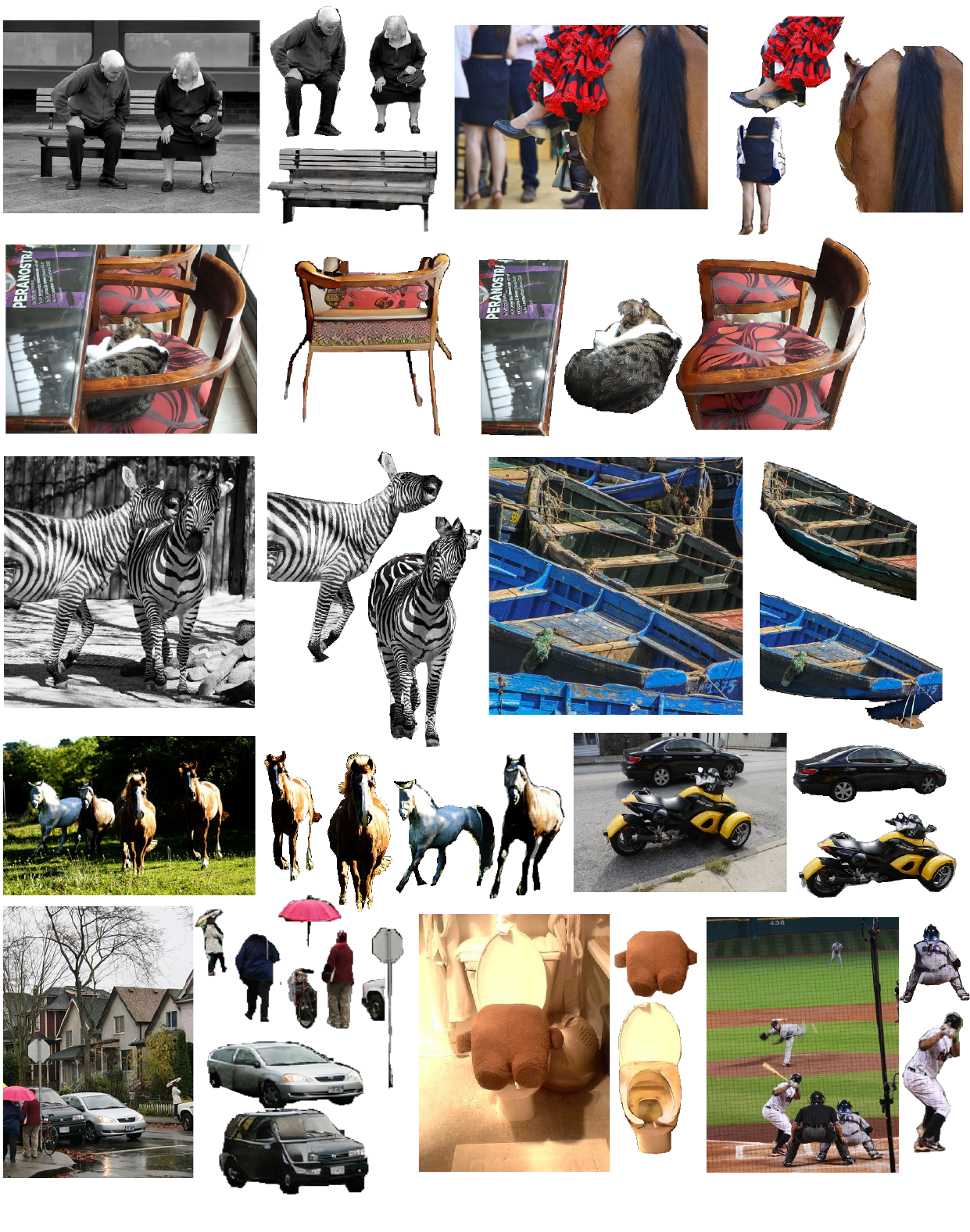}
\caption{Additional deocclusion results of PACO in real-world scenes. 
}
\label{fig:4}
\vspace*{-2.75mm}
\end{figure*}

\begin{figure*}
\centering
\includegraphics[width=0.99\textwidth]{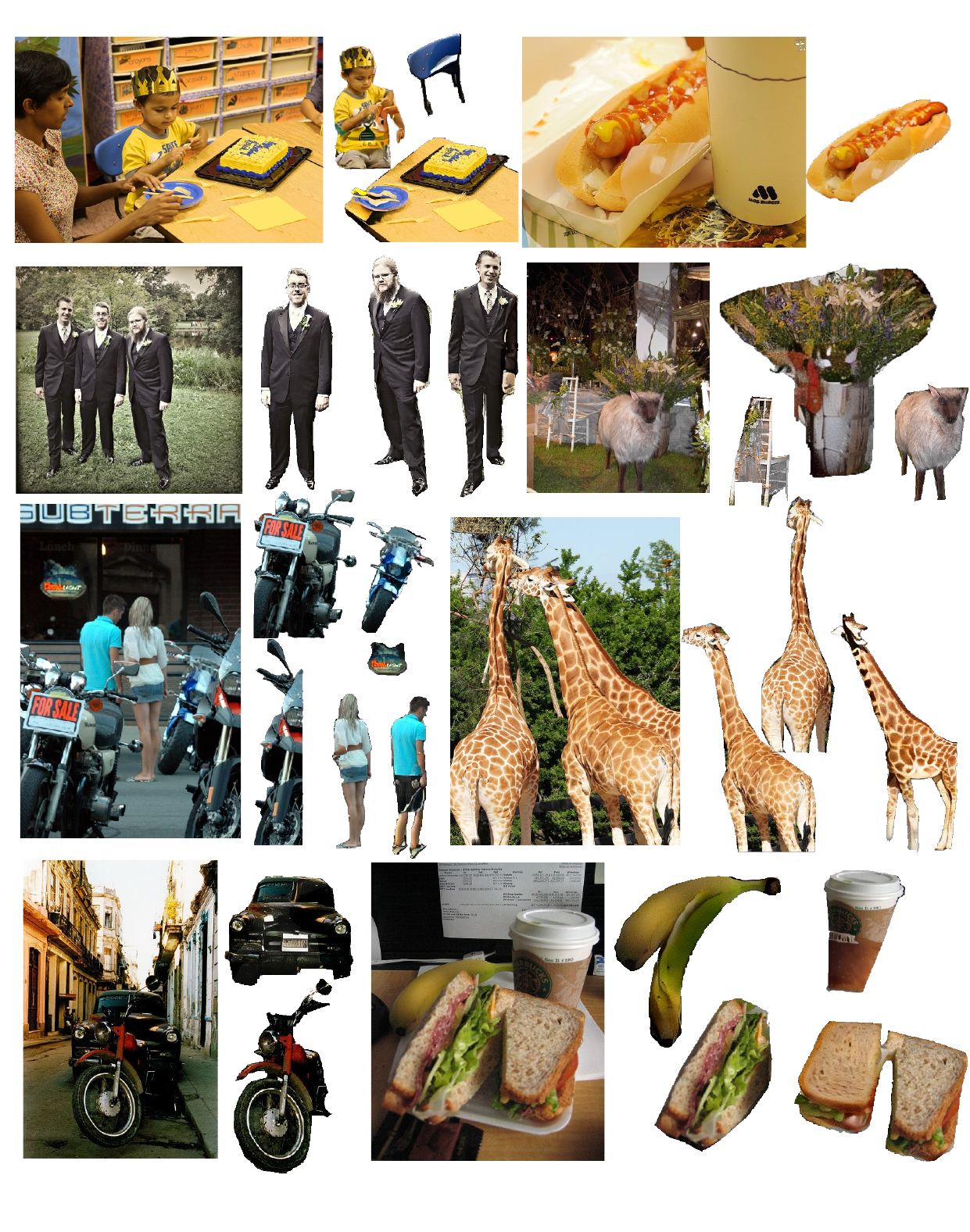}
\caption{Additional deocclusion results of PACO in real-world scenes.
}
\label{fig:5}
\vspace*{-2.75mm}
\end{figure*}

\begin{figure*}
\centering
\includegraphics[width=0.99\textwidth]{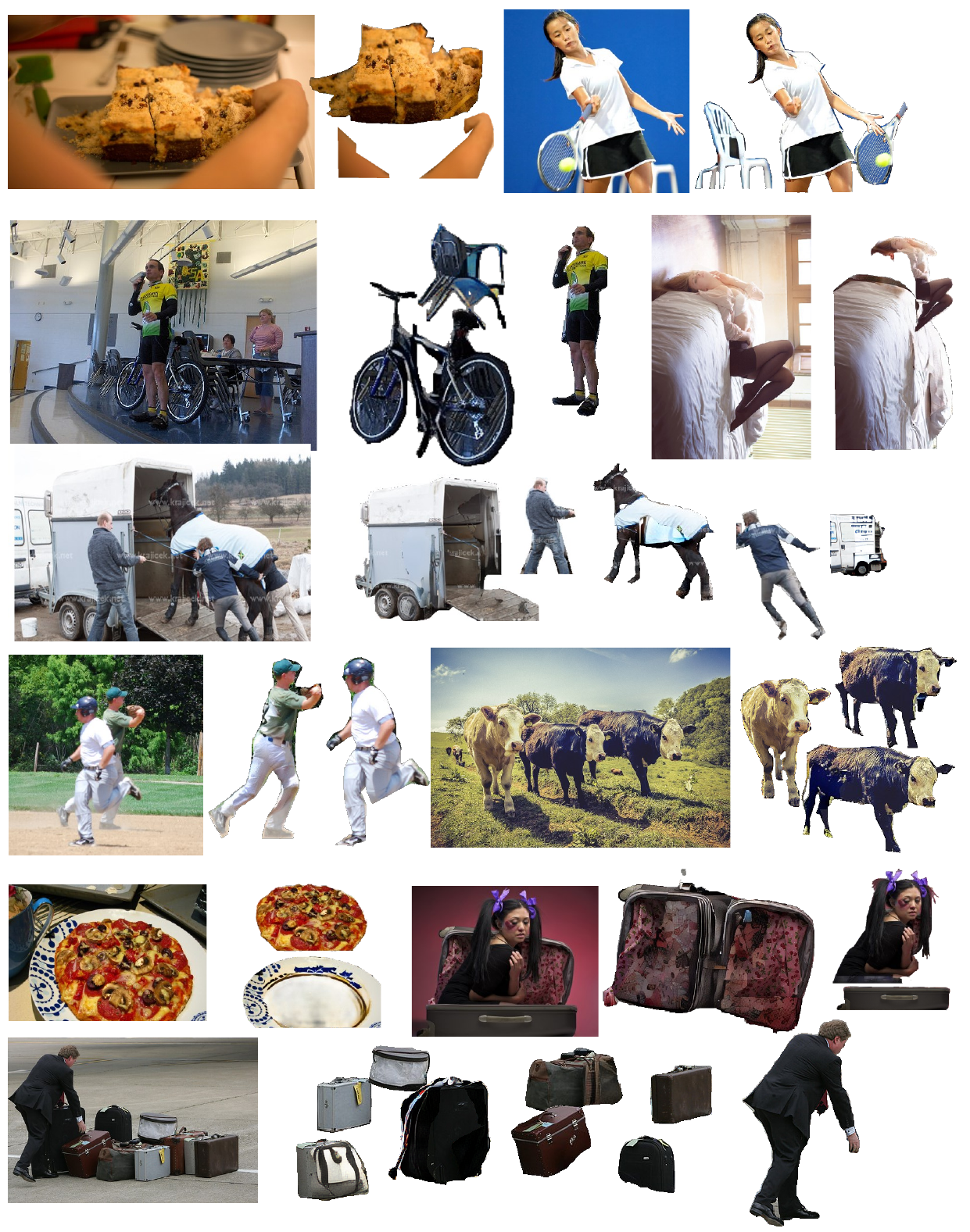}
\caption{Additional deocclusion results of PACO in real-world scenes.
}
\label{fig:6}
\vspace*{-2.75mm}
\end{figure*}

\begin{figure*}
\centering
\includegraphics[width=0.99\textwidth]{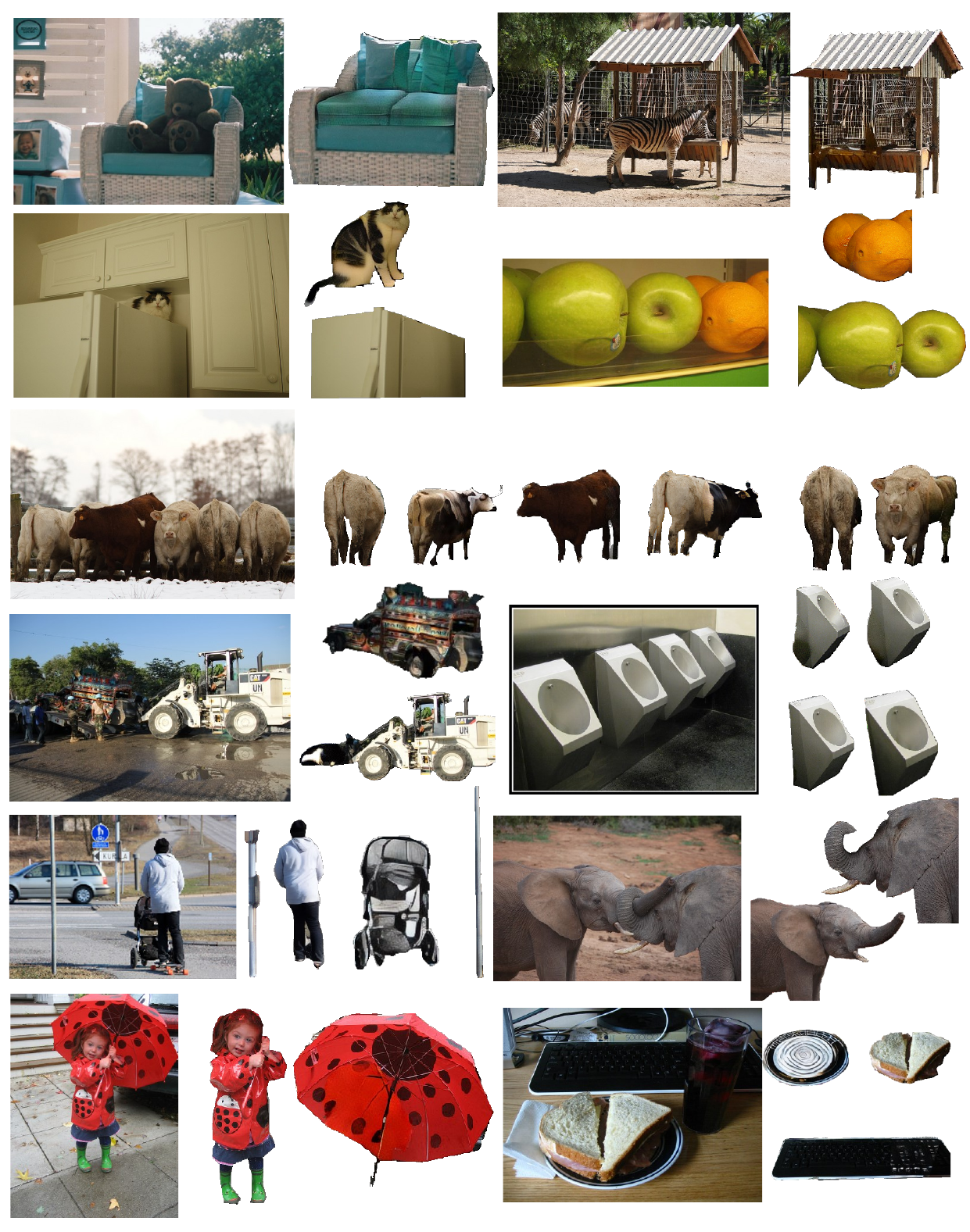}
\caption{Additional deocclusion results of PACO in real-world scenes.
}
\label{fig:7}
\vspace*{-2.75mm}
\end{figure*}

\begin{figure*}
\centering
\includegraphics[width=0.99\textwidth]{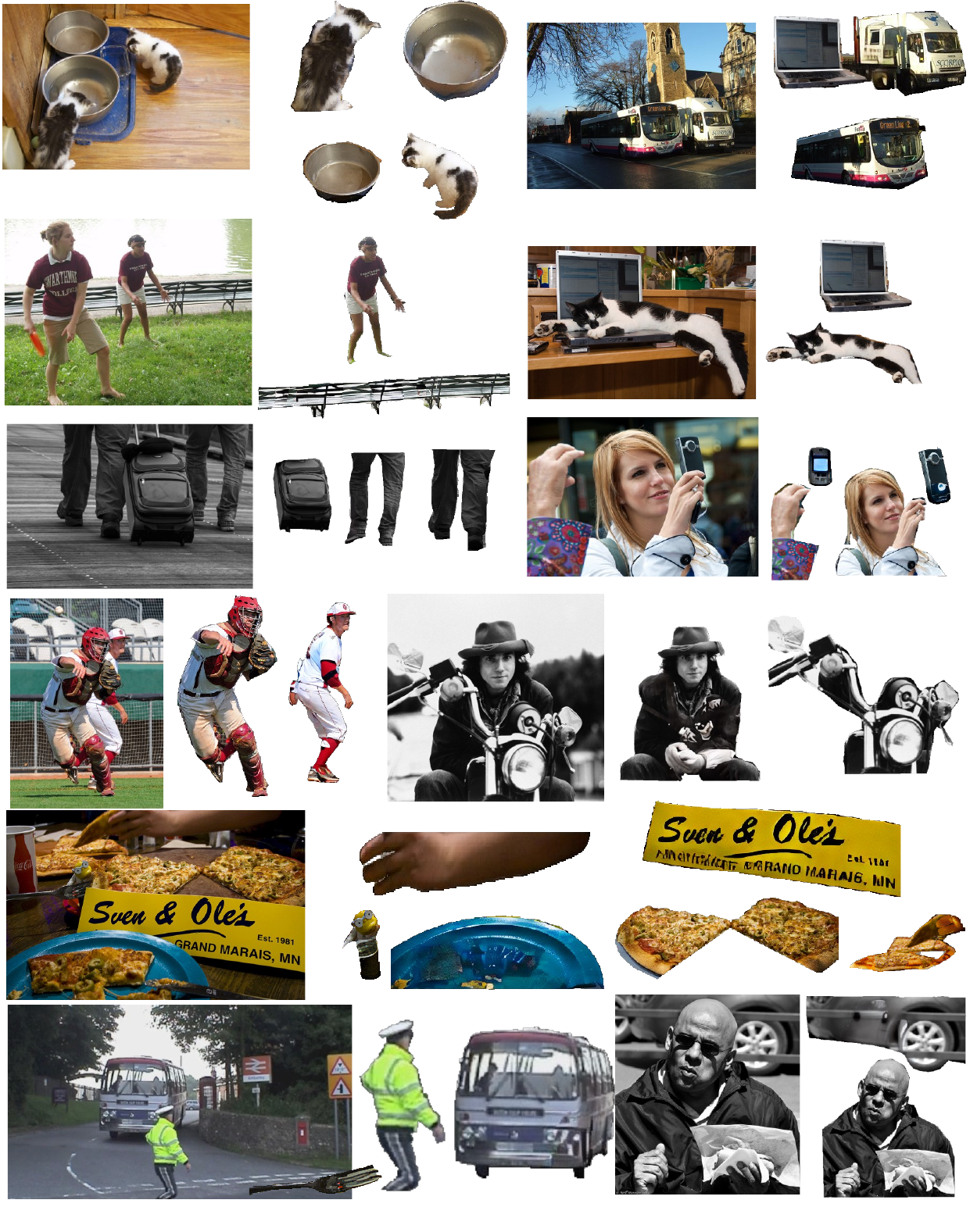}
\caption{Additional deocclusion results of PACO in real-world scenes.
}
\label{fig:8}
\vspace*{-2.75mm}
\end{figure*}

\section{Limitations}

While our approach shows superior performance, it still has certain limitations:


(i) \textbf{Handling shadows on objects.} Our method faces challenges with shadows. For example, in the bottom right image of Figure~\ref{fig:3} in the supplementary file, shadows cast on the skiboard create black artifacts. Additionally, shadows in the background can impact the consistency of recomposed scenes, making shadow handling a crucial area for future research.

(ii)  \textbf{Dependency on high-quality segment maps and text annotations.} Our process requires quality segment maps and text annotations for each object, Although they can be acquired with foundation models like SAM~\cite{kirillov2023segment} and GPT-4V~\cite{openai2023gpt4v}, respectively, this dependency can hinder full automation for processing in-the-wild images. In addition, our performance is limited by the quality of the segmentation map. 

(iii) \textbf{Limitations with objects partially exceeding the image boundary.} Our current system struggles with objects that extend beyond the image boundary. For instance, in the top right image of Figure~\ref{fig:1}, the left side of a teddy bear is not completed as it crosses the image's boundary. One solution to complete this bear is to pad a border with the proper size on the top left boundary of the image; yet, it requires additional pre-processing by the user.

(iv) \textbf{Background deocclusion is limited by the inpainting models.} Since background inpainting has been extensively studied, our approach is mainly designed for object-level deocclusion, leaving background processing to inpainting models like LAMA~\cite{suvorov2022resolution}. Hence, the quality of the background in our recomposed images is limited by these inpainting models' capabilities. Deocclusion on both the objects and background can be a future direction. 

(v) \textbf{Lack of 3D awareness.} The absence of 3D awareness in our method can lead to unrealistic deocclusion, as seen in the bottom right example of Figure~\ref{fig:5}, where a banana and two sandwiches unrealistically share the same 3D space. Incorporating 3D awareness into object-level scene deocclusion is a promising direction for future exploration.

\bibliographystyle{ACM-Reference-Format}
\bibliography{bibliography}

\end{document}